\documentclass[journal]{IEEEtran}

\usepackage{moreverb,url}
\usepackage[colorlinks,bookmarksopen,bookmarksnumbered,citecolor=red,urlcolor=red]{hyperref}

\usepackage{amsmath,amsfonts,amssymb}
\usepackage{algorithm}
\usepackage[noend]{algpseudocode}
\usepackage{array}
\usepackage[caption=false,font=normalsize,labelfont={sf,scriptsize},textfont={sf,scriptsize}]{subfig}
\usepackage{textcomp}
\usepackage{verbatim}
\usepackage{graphicx}
\usepackage{enumitem}
\usepackage{tikz}

\usepackage[font=footnotesize]{caption}

\usepackage{xargs}
\usepackage[table]{xcolor}
\usepackage{color}
\usepackage{listings}
\usepackage{lstlang-ampl}

\usepackage{bm}
\usepackage{xparse}
\usepackage{siunitx}

\usepackage{multirow}
\usepackage{booktabs}

\usetikzlibrary{
  arrows.meta, 
  calc, 
  3d, 
  positioning, 
  decorations.pathreplacing,
  decorations.pathmorphing
}
\usepackage{pgfplots}
\pgfplotsset{compat=1.18}
\usepgfplotslibrary{statistics}
\usepackage{pgfplotstable}

\graphicspath{{fig/}}

\newcommand{\Lk}[1]{\os{L}_{#1}}

\newcommand{\os}[1]{\mathrm{#1}}
\NewDocumentCommand{\osi}{m O{i}}{\os{#1}_{#2}}

\NewDocumentCommand{\gosi}{m O{i}}{\os{#1}^{#2}}

\newcommand{\us}[1]{\mathcal{#1}}
\newcommand{\usi}[1]{\mathcal{#1}^i}

\newcommand{\cs}[1]{\mathcal{#1}_\text{con}}
\NewDocumentCommand{\csi}{m O{i}}{\cs{#1}^{#2}}

\newcommand{\Acon}{\mathrm{A}_\text{con}^i}

\NewDocumentCommand{\con}{m O{}}{\spat[#2]{\mathfrak{#1}}[ik]}

\DeclareMathOperator{\pos}{position}
\DeclareMathOperator{\ori}{orientation}

\newcommand{\pd}[2]{{\frac{\partial #1}{\partial #2}}}
\newcommand{\sd}[2]{{\frac{d #1}{d #2}}}

\newcommand{\R}{\mathbb{R}}

\newcommand{\fun}[3]{#1: #2 \to #3}

\NewDocumentCommand{\spat}{o t\dot m o o}{%
    \IfValueTF{#1}{%
        {^{#1}}%
    }{%
        {}%
    }%
    \IfBooleanTF{#2}{%
        \dot{\bm{#3}}%
    }{%
        \bm{#3}
    }%
    \IfValueTF{#5}{%
        \IfValueTF{#4}{%
            {_{#4}^{#5}}%
        }{%
            {^{#5}}%
        }%
    }{%
        \IfValueTF{#4}{%
            {_{#4}}%
        }{%
            {}%
        }%
    }%
}

\newcommand{\spatv}{\spat{v}}
\newcommand{\spata}{\spat{a}}
\newcommand{\spatf}{\spat{f}}
\newcommand{\spats}{\spat{s}}
\newcommand{\spatI}{\spat{I}}
\newcommand{\spatr}{\spat{r}}
\newcommand{\spatX}{\spat{X}}

\newcommand{\spatdot}[3]{{^{#1}\dot{\spat{#2}}_{#3}}}

\newtheorem{myrem}{Remark}
\newtheorem{mydef}{Definition}

\algnewcommand{\ForOne}[1]{\State\algorithmicfor\ #1\ \algorithmicdo}
\algnewcommand{\ConOne}[1]{\State\textbf{subject to}\ #1}
\algblockdefx[CON]{Con}{EndCon}[1][]{\textbf{subject to} #1}{}
\makeatletter
\ifthenelse{\equal{\ALG@noend}{t}}%
  {\algtext*{EndCon}}
  {}%
\makeatother

\begin{document}
\title{Amplify: A Lightweight Library for Reproducible Nonlinear Programming Problems in Robotics}
\author{Nelson Rosa Jr.\thanks{email: nr@u.northwestern.edu.}}
\maketitle

\markboth{}%
{Rosa: A Lightweight Library for Reproducible NLPs in Robotics}

\begin{abstract}
Optimization problems (OPs) are key to solving many challenging research
problems in robotics.  However, reproducibility still remains a major issue.  In
this paper, we present Amplify, a lightweight nonlinear programming library
aimed at reproducible results of robotic-related trajectory optimization
problems.  The minimalistic requirements for the 537-line library (80 characters
per line) are an Internet connection, familiarity with the AMPL modeling
language, and a text editor.  Our primary contribution is the formulation of a
library where trajectory optimization algorithms are represented directly within
the optimization model.  Specifically, we implement the algorithms used to
compute the dynamics, trajectories, and reference motions as constraints of the
OP in a declarative programming paradigm.  We outline how our formulation of
objectives, decisions variables, and constraints can be implemented in other
transcription libraries that want to be lightweight and reproducible.  We also
compare the Amplify framework with 3 other libraries across examples of
benchmark optimization problems across several fields, including bipedal
locomotion and grasp planning.
\end{abstract}

\section{Introduction}
\IEEEPARstart{T}{rajectory} optimization problems form a core set of problem types in robotics.
The numerical computation of an optimal solution often requires transcribing the
problem into a nonlinear programming problem (NLP).  While there exists many
frameworks that attempt to simplify the challenges of transcribing optimization
problems into NLPs \cite{Fevre2020, Ruscelli2022, Vanroye2023a, Howell2019,
Kelly2017, Mastalli2020, Jallet2025, Tedrake2019}, end-users wanting to
reproduce the work often cannot.  Two common problems are 1) accessibility to
the optimization solvers and 2) the software toolchain necessary for computing
the optimal solutions.  Developers also face challenges as customizing the
codebase requires careful navigation of heavyweight APIs in order to apply the
library to their particular use case.

In this paper, we present Amplify. Amplify is a nonlinear programming (NLP)
library for solving hybrid trajectory optimization problems (OPs). The framework
is designed to be self-contained (requires no external libraries), reproducible
(at most 3 files are required to replicate a solution), and accessible (the
solvers and software toolchain are available to anyone with an Internet
connection).

The core design decisions that lead to these features are not limited to a
particular library stack.  In our work, we use AMPL \cite{Fourer1990,
Fourer2011, Fourer2004} as the modeling language for demonstration purposes.
This choice has the additional benefit of leveraging the Network-Enabled
Optimization Server (NEOS) for solving NLPs \cite{Czyzyk1998}.  NEOS is a free
cloud-based service with access to 32 commercial and open-source solvers that
accept AMPL as input.  The server allows users to run large-scale optimization
problems of up to 8~GB in RAM for up to 8~hours on an HTC Condor computing grid.
The use of NEOS for solving robotic optimization problems keeps our framework
accessible to users with an Internet connection \cite{Czyzyk1999}.

As a technical contribution to addressing challenges in reproducibility, our
framework implements algorithms as constraints in the NLP of the three-most
common tasks of a transcription library \cite{Betts2020}: the 1) system
dynamics, 2) ODE solver, and 3) polynomial reference trajectories.  Embedding
the transcription library into the NLP removes a key hurdle to reproducibility
prominent in other libraries.  With other libraries the NLP (i.e., the
collection of objectives, decision variables, and constraints) is generated at
runtime as an output.  This common design decision forces the user to rely on
having the library's toolchain to compute the results.  This often leads to
issues with reproducibility.  This paper presents a reference implementation to
overcome this limitation.  In our work, there is a a clean separation of the
mathematical model (a generic description of the problem) from the data (the
specific instance of the problem to solve).  This creates a self-contained NLP
model file that does not have to be regenerated every time any of the three
tasks changes.  With Amplify, the model is the library.

\subsection{Previous Work} 
The inability of researchers to reproduce published work is a major problem
across many scientific fields, including robotics \cite{Bonsignorio2025,
Cervera2024, Cervera2019}.  The extent of the problem in the
optimization-related robotics literature has yet to be formally documented, but
anecdotally exists.  In several instances, the issue is in how the OP is
implemented in software.

The general approach in the literature is to directly implement an OP at the
API-level of the solver (the numerical optimization software)
\cite{Bessonnet2005, Griffin2015a}.  For example, commonly used solvers in the
bipedal walking literature are SNOPT, IPOPT, and fmincon \cite{Gill2002,
Waechter2005, MathWorks2025}.  For widespread dissemination that encourages
further research, this approach is suboptimal.

Mixing the optimization problem with the solution method is suboptimal when
compared to how the operations research community has for decades benefited from
separating the description of an optimization problem from the solver
\cite{Jusevicius2021}.  This separation is commonly achieved through the use
of mathematical modeling languages, such as AIMMS, AMPL, GAMS, JModelica, JuMP,
Pyomo, and CVX \cite{Dunning2017, Hart2011}.

In particular, algebraic modeling languages (AML) \cite{Jusevicius2021}, like
AIMMS, AMPL and GAMS, are declarative languages widely used in industry and
academia.  They represent optimization problems using sets, parameters,
variables, objectives, and constraints.  An AML compiler also provides useful
services such as targeting multiple solvers, simplifying the problem instance by
tightening bounds and eliminating unused variables, and deriving first- and
second-order derivatives \cite{Fourer1994, Gay1996}.

While these modeling languages can express a broad range of nonlinear
programming problems (NLPs), they are not well suited for directly expressing
trajectory optimization problems in robotics (ROPs) as NLPs.  Most roboticists
would find it burdensome to, for example, code a robot's equations of motion by
hand when standards like URDF files exist to offload the burden to the
application.  This has led to robotics-focused libraries that automate many of
the standard tasks of expressing the ROP as an NLP (e.g., formulating the
equations of motion, reference trajectories, and integration schemes as
constraints).  Table~\ref{tab:rtops} shows a list of libraries we've surveyed
for this paper.

When surveying the programming languages and environments used by ROP
libraries, we see that library developers in our survey do not build their
libraries on top of modeling languages.  The trend is to implement ROPs in a
general-purpose programming language (GPL) using an object-oriented programming
(OOP) approach.  A few of the libraries, like Drake, further limit themselves to
variants of *nix operating systems (i.e., Linux and MacOS).

In terms of dependencies, several libraries rely on CasADi to compute
derivatives.  Computing derivatives as a built-in feature for a ROP library is
useful as many libraries target gradient-based solvers.  As an alternative to
CasADi, PyRobotCop uses ADOL-C.  Examples of popular general-purpose
gradient-based solvers are IPOPT, qpOASES, and SNOPT.  Custom solvers that focus
on ROPs are FATROP, Altro, and custom Gauss-Newton QP solvers.
\begin{table*} 
\centering 
\scriptsize
\begin{tabular}{c|c|c|c|c}
Library & Languages [Add'l Bindings] & User API & Solvers & Lines of Code [\# of
Files] \\ \hline 
Amplify & AMPL [python, R, C\#, C++, Java, Matlab] & Declarative & IPOPT and 25+ others & \\
acado & C++ [Matlab] & Object Oriented & qpOASES and 3+ others & 271877 [1968] \\ 
acados & C [python, Matlab, Octave] & Object Oriented & qpOASES and 9+ others & 787963 [1191] \\ 
CasADi's Opti & C++ [python, Matlab, Octave] & Object Oriented & IPOPT and 21+ others & 334875 [1496] \\ 
do-mpc & python & Object Oriented & IPOPT & 30438 [208] \\ 
Drake & C++ [python] & Object Oriented & IPOPT and 12 others & 842197 [5172] \\ 
FROST & Matlab & Object Oriented & IPOPT & 99725 [1357] \\ 
Horizon & python & Object Oriented & custom, IPOPT & 49130 [300] \\ 
OptimTraj & Matlab & Procedural & fmincon & 18508 [189] \\
PyRoboCop & python & Object Oriented & IPOPT & 49974 [141] \\ 
roboptim & C++ & Object Oriented & IPOPT and 3 others & 19816 [223] \\
ROCKIT & python [Matlab] & Object Oriented & IPOPT & 16770 [146] \\
TACO & AMPL & Declarative & MUSCOD-II & 39728 [141] \\ 
TrajectoryOptimization.jl & Julia & Object Oriented & Altro & 6185 [68] \\ 
TRAJOPT & C++ & Object Oriented & qpOASES and 3 others & 86360 [456] \\ 
trep & python & Object Oriented & custom & 23236 [188] \\ 
TROPIC & Matlab & Object Oriented & IPOPT & 1309945 [2280] \\
\end{tabular}
\caption{A list of trajectory optimization libraries.  The
languages and API are from the user's programming perspective.  The lines of
source code (and files) are taken for their respective online repositories as
cloned on Febuary 28, 2026 after running \texttt{CLOC} \cite{Danial2021}.}
\label{tab:rtops}
\end{table*}

Another common feature is the need to symbolically compute the robot dynamics.
The growing use of Pinocchio in ROP software stacks, a rigid-body dynamics
software library, underscores the fact that trajectory optimization problems
require the computation of a robot's equations of motion in terms of the
decision variables and parameters of the ROP.  Other rigid-body dynamics
libraries used are RBDL and \verb+spat_v2+ \cite{Featherstone2012, Felis2016}.
Standard rigid-body dynamics algorithms used to compute the equations of motion
include the Recursive Newton-Euler (RNEA), Composite Rigid Body (CRBA), and
Articulated Body algorithms.  These are linear-time algorithms used to compute
generalized forces, mass matrices, and generalized accelerations, respectively.
Library developers make heavy use of spatial vectors to achieve efficient
implementations of these algorithms \cite{Featherstone2008, Lynch2023}.

While tasks like computing derivatives and rigid-body dynamics are often left to
external libraries, all of the ROP libraries in Table~\ref{tab:rtops} implement
their own code for declaring decision variables, objectives, and constraints.
This is in spite of the ubiquitous presence of CasADi, which has its own NLP
modeling interface (i.e., Opti), and many modeling languages, like AMPL, that
provide software bindings to other languages.

In the end, what ROP libraries do well is separate the optimization problem
from the solver.  This is arguably their biggest weakness when it comes to
reproducibility.  In order for an end-user to replicate the optimal trajectory,
they will need the problem formulation written using the ROP's API, ROP
library, GPL, and any software dependencies to replicate the result.  Extending
the NLP also requires the entire toolchain, even if the change is trivial.

When the toolchain works, the application of ROPs to solve challenging motion
tasks in robotics is well documented.  Application areas include legged
locomotion \cite{Wensing2024}, manipulation, unmanned vehicles, and medical
robotics.  Successful examples of deployment on experimental platforms include
the biped walkers RABBIT, Mabel, Marlo, and Cassie.  These bipeds use the hybrid
zero dynamics framework to solve ROPs that find periodic trajectories and
closed-loop control laws for stabilizing the actuated motions of the
trajectories.  Libraries dedicated to solving this particular type of ROP are
TROPIC and FROST \cite{Fevre2020, Hereid2018}.

For comparing ROP libraries, the hybrid zero dynamics framework provides a
useful benchmark for optimization problems of hybrid dynamical systems.  The
framework defines a parameterized control law through a set of virtual
constraints (constraints on the system that are enforced through feedback
control) \cite{Westervelt2007}.  These constraints are typically reference
trajectories that the actuated joints of the robot have to track
\cite{Ramezani2013}.  Models of constrained hybrid dynamical systems are
commonplace in legged locomotion and manipulation.  Providing features for
modeling these types of dynamical systems in ROPs is a must in these areas of
robotics.


\subsection{Statement of Contributions} 
\label{ssec:soc}
Amplify is a robotics-focused trajectory optimization library.  It can model
hybrid dynamical systems with physical and virtual constraints, integrate the
equations of motion using user-defined Runge-Kutta methods, specify B\'{e}zier
curves for joints and control inputs to track, and be extended with user-defined
objectives, decision variables, and constraints.

Overall, our work presents an example of how ROPs can be built on top of a
modeling language, and demonstrate the benefits of such a design decision.  In
general, our focus is not only in promoting Amplify, but potentially modernizing
how ROP libraries are developed in the future.  Relative to the state of the
art, we provide an alternative view of what it means to be modular,
reproducible, and lightweight.  Specifically, our contributions are
\begin{enumerate}

\item \textbf{An optimization library paradigm for reproducible robotics
research.}  We demonstrate the implementation of an NLP library that has minimal
dependencies (an Internet connection and a text editor), emphasizes
reproducibility (at most 3 text files need to be distributed to reproduce a
result), and is accessible (the dependencies are freely available in many parts
of the world).  Our approach is generic and can be implemented in other
optimization frameworks.

\item \textbf{The representation of computationally efficient rigid-body
dynamics algorithms as optimization constraints.}  We eliminate a major library
dependency by rewriting the Recursive Newton-Euler, Composite Rigid Body, and
Operational Space Inertia Matrix algorithms as constraints in an NLP.  We
demonstrate how these algorithms and their data structures can be used to
compute optimal trajectories of constrained hybrid dynamical systems.

\item \textbf{Generic implementations of Runge-Kutta methods and B\'{e}zier
curves as optimization constraints.}  We demonstrate how libraries can implement
arbitrary integration schemes and polynomial curves (e.g., curves for a
manipulator's end effector or mobile robot to track) as optimization
constraints.  Our implementation permits greater expressiveness in how the NLP
is formulated relative to standard ROP implementations, which have a small
number of pre-defined options.

\item \textbf{A unified approach to modeling physical and virtual constraints of
a dynamical system.}  Our modeling approach takes advantage of the similar
computations between physical and virtual constraints.  Furthermore, we aim to
be the library of choice for reproducible results for researchers who rely on
virtual constraints in their robot models (e.g., the hybrid zero dynamics
framework).

\item \textbf{The use of the Network-Enabled Optimization System (NEOS) for
reproducing published work.}  To our knowledge, we are the first to utilize this
cloud-based service as a viable tool for solving optimization problems in
robotics.  The cloud-based NEOS server provides access to 32 open-source and
commercial solvers that accept AMPL input for free.  This provides a valuable
prototyping tool.

\item \textbf{A baseline for optimization libraries in robotics.}  Given the
range of solvers our library can target, reproducibility of the results, and
range of problems we can tackle, other optimization libraries can compare their
implementations and results to our library.

\end{enumerate}
Amplify and associated tools are available at
\url{https://github.com/nr-codes/Amplify}.

\subsection{Paper Outline}
\begin{figure*}[t]
\centering
\begin{tikzpicture}
  \tikzset{every node/.style={font=\scriptsize}}

  \node (G) [shape=rectangle,draw,anchor=north west] at (0,9) {
    \begin{tabular}{@{}l@{}}
      \multicolumn{1}{c}{\textbf{Grid}} \\ \hline \\[-1.5ex]
      sets: $\us{S}$, $\us{P}$, $(\gosi{G}, \leq)$, $(\gosi{G}[0], \leq)$ \\
      declares: $t$ (discretized time)
    \end{tabular}
  };

  \node (B) [shape=rectangle,draw,anchor=north west] at ($(G.north east)+(1cm,0)$) {
    \begin{tabular}{@{}l@{}}
      \multicolumn{1}{c}{\textbf{B\'{e}zier}} \\ \hline \\[-1.5ex]
      constraints: Equation~\ref{eqn:bezd} \\
      declares: $P_i$, $\tau_i$ of Equation~\ref{eqn:bezd}
    \end{tabular}
  };

  \node (KT) [shape=rectangle,draw, anchor=north west] at ($(B.north east)+(1cm,0)$) {
    \begin{tabular}{@{}l@{}}
      \multicolumn{1}{c}{\textbf{Robot Kinematic Tree}} \\ \hline \\[-1.5ex]
      sets: $L_0$, $L_1$, $L_{-1}$, $(\osi{P}, \leq)$
    \end{tabular}
  };

  \node (S) [shape=rectangle,draw,below=2cm of G.center] {
    \begin{tabular}{@{}l@{}}
      \multicolumn{1}{c}{\textbf{State}} \\ \hline \\[-1.5ex]
      sets: $Q$, $X$ \\
      declares: $q$, $\dot{q}$, $\ddot{q}$, $u$
    \end{tabular}
  };

  \node (SQ) [shape=rectangle,draw] at ($(B.center |- S.center)$) {
    \begin{tabular}{@{}l@{}}
      \multicolumn{1}{c}{\textbf{Spatial Quantities}} \\ \hline \\[-1.5ex]
      constraints: Section~\ref{sec:sv} \\
      defines: $\spatX$, $\spats$, $\spatI$, $\spata_g$
    \end{tabular}
  };

  \node (C) [shape=rectangle,draw] at ($(KT.center |- SQ.center)$) {
    \begin{tabular}{@{}l@{}}
      \multicolumn{1}{c}{\textbf{Robot Constraints}} \\ \hline \\[-1.5ex]
      sets: $R$, $\usi{R}_r$, $\csi{J}$, $(\Acon, \leq)$
    \end{tabular}
  };

  \node (CRB) [shape=rectangle,draw,below=2cm of S.center] {
    \begin{tabular}{@{}l@{}}
      \multicolumn{1}{c}{\textbf{Composite Rigid Body}} \\ \hline \\[-1.5ex]
      constraints: Algorithm~\ref{alg:crba} \\
      defines: $M$ of Equation~\ref{eqn:M}
    \end{tabular}
  };

  \node (RNEA) [shape=rectangle,draw] at ($(SQ.center |- CRB.center)$) {
    \begin{tabular}{@{}l@{}}
      \multicolumn{1}{c}{\textbf{Recursive Newton-Euler}} \\ \hline \\[-1.5ex]
      constraints: Algorithm~\ref{alg:rnea} \\
      defines: $b$ of Equation~\ref{eqn:M}
    \end{tabular}
  };

  \node (TJA) [shape=rectangle,draw] at ($(C.center |- RNEA.center)$) {
    \begin{tabular}{@{}l@{}}
      \multicolumn{1}{c}{\textbf{Task Jacobian}} \\ \hline \\[-1.5ex]
      constraints: Algorithm~\ref{alg:tja} \\
      defines: $J^i$, $\phi^i$, $B^i$, $y^i$, $\dot{y}^i$ \\
      \qquad of Equations~\ref{eqn:M}, \ref{eqn:I}, \ref{eqn:v}
    \end{tabular}
  };

  \node (EOM) [shape=rectangle,draw, below=2cm of RNEA.center] {
    \begin{tabular}{@{}l@{}}
      \multicolumn{1}{c}{\textbf{Equations of Motion}} \\ \hline \\[-1.5ex]
      constraints: Equations~\ref{eqn:M}--\ref{eqn:I} \\
      declares: $f$, $\iota$ of Equations~\ref{eqn:M}--\ref{eqn:I}
    \end{tabular}
  };

  \node (RK) [shape=rectangle,draw] at ($(TJA.center |- EOM.center)$) {
    \begin{tabular}{@{}l@{}}
      \multicolumn{1}{c}{\textbf{Runge-Kutta}} \\ \hline \\[-1.5ex]
      constraints: Section~\ref{sec:ode} \\
      declares: $h$ of Equation~\ref{eq:RK}
    \end{tabular}
  };

  \path (G.east) edge[<-] (B.west |- G.east);
  \path (G.south) edge[<-] (G.south |- S.north);
  \path (KT.south west) edge[<-] (SQ.north east);
  \path (S.east) edge[<-] (SQ.west |- S.east);
  \path (SQ.east) edge[->] (C.west |- SQ.east);
  \path (SQ.south west) edge[<-] (CRB.north east);
  \path (SQ.south) edge[<-] (SQ.south |- RNEA.north);
  \path (SQ.south east) edge[<-] (TJA.north west);
  \path (CRB.south east) edge[<-] (EOM.north west);
  \path (RNEA.south) edge[<-] (RNEA.south |- EOM.north);
  \path (TJA.south west) edge[<-] (EOM.north east);
  \path (EOM.east) edge[<-] (RK.west |- EOM.east);

\end{tikzpicture}
\caption{The modules of the Amplify library.  The sets, constraints, and
variables listed in each block are defined throughout the paper.  They represent
a sample of the content defined within a module.  Arrows between blocks denotes
a dependency.  For example, the B\'{e}zier module depends on the content defined
in the Grid module.  In AMPL, decision variables are declared (e.g., $f$ and
$\iota$ in the \textbf{Equations of Motion} module).  On the other hand, defined
variables are variables expressed in terms of other variables (e.g., $M$ in the
\textbf{Composite Rigid Body} module depends on $q$ and several other variables).}

\label{fig:mods}
\end{figure*}
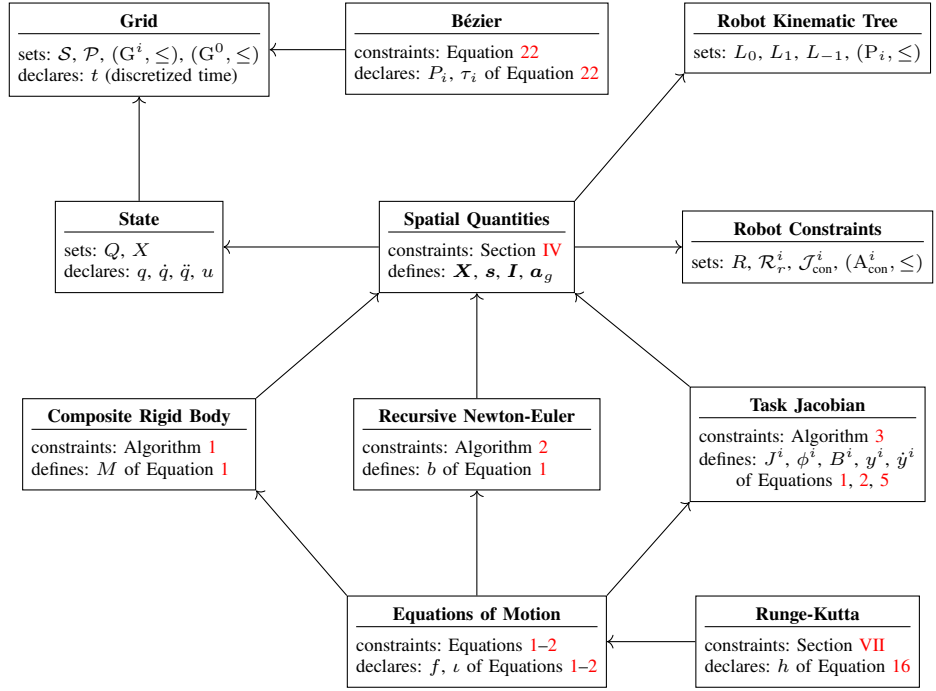
Given our contributions, this paper targets 3 distinct audiences:
\begin{enumerate}
\item developers of (optimization) modeling languages not familiar with
robotics, but hoping to extend their libraries to accommodate the needs of a
large swath of the field,
\item developers of robotic transcription libraries who want their libraries to
shed a few pounds (i.e., remove dependencies to become more lightweight), and
\item users who want a straightforward and reproducible optimization problems
that they can run and extend as needed.
\end{enumerate}

To accommodate these different target audiences, our goal is to present concepts
and algorithms with sufficient details that others can implement our
contributions into their own codebase.  In the remainder of this paper, we
introduce the equations of motion for the class of hybrid dynamical systems that
we model in Sections~\ref{sec:H}--\ref{sec:rc}.  Section~\ref{sec:sv} details
how spatial vectors are used to compute the various quantities in the equations
of motion.  Sections~\ref{sec:grid}--\ref{sec:bez} document the implementation
of the system dynamics, ODE solver, and polynomial reference trajectory modules
in the Amplify framework.  The framework is then compared against benchmark
problems and libraries found across the field of robotics in
Section~\ref{sec:ex}.  We discuss the results and conclude in
Sections~\ref{sec:dis}--\ref{sec:con}, respectively.  Additionally,
Figure~\ref{fig:mods} presents a visual overview of how the hybrid dynamics and
the core Amplify modules relate to each other.


\section{Modeling Robot Trajectories}
\label{sec:H}
We consider a system of $N$ robots as an impulsive dynamical system with
$m$ phases of locomotion.  The continuous dynamics of each phase is
modeled as a constrained mechanical system:
\begin{equation}
\begin{gathered}
M(q)\ddot{q} + b(q, \dot{q}) = u + B^i(q) f^i \\
J^i(q) \ddot{q} + \dot{J}^i(q) \dot{q} + \phi^i(q, \dot{q}, t) = 0
\end{gathered} \quad 1 \leq i \leq m,
\label{eqn:M}
\end{equation}
where $q \in \R^{n_q}$ are the generalized coordinates of the robots; $M(q) \in
\R^{n_q \times n_q}$ is the system's mass matrix; $b(q, \dot{q}) \in \R^{n_q}$
is the force due to gravity, Coriolis, and centrifugal terms; $u \in \R^{n_q}$
is the control input; $f^i \in \R^{n_{f^i}}$ are constraint forces and inputs
(see Section~\ref{ssec:uni} for further details); $B^i(q) \in \R^{n_q} \times
\R^{n_{f^i}}$ is the transmission matrix; $J^i(q), \dot{J}^i(q) \in \R^{n_{J^i}
\times n_q}$ are the constraint Jacobian and its time derivative, respectively;
and $\phi^i(q, \dot{q}, t) \in \R^{n_{J^i}}$ is a vector of additional
constraints.

In general, superscripts on the rightside of a variable denotes that its
definition and dimension are phase dependent.  No superscript implies there is
no phase dependency.  For example, the terms $M$, $b$, and $u$ form the
unconstrained dynamics of the system.  These quantities are active across all
time and their dimensions do not change across phases.  The quantities $J^i$,
$B^i$, $\phi^i$, and $f^i$ define the set of constraints and constraint forces
active during the $i^\text{th}$ phase.  The dimension of these quantities can
change across phases.

If there is an impact across two phases, say $i$ and $j$, we use the
impulse-momentum equation to compute the pre- and post-impact velocities:
\begin{equation}
\begin{aligned}
q^+ &= q^- \; (= q) \\
M(q)(\dot{q}^+ - \dot{q}^-) &= \left(J^i(q)\right)^T \iota^i \\
J^i(q)\dot{q}^+ &= E^{ij} J^j(q)\dot{q}^-,
\end{aligned}
\label{eqn:I}
\end{equation}
where $+$ and $-$ denote post- and pre-impact quantities, respectively, $\iota
\in \R^{n_{\iota^i}}$ is an impulse, and $E^{ij} \in \R^{n_{J^i} \times
n_{J^j}}$ is a matrix of restitution coefficients relating the pre-impact
velocity $\dot{q}^-$ in phase $j$ to the post-impact velocity $\dot{q}^+$ in
phase $i$.

Together \eqref{eqn:M} and \eqref{eqn:I} define the robot's hybrid dynamics
during phase $i$ and transition into phase $j$.  Collectively, these form the
system's hybrid dynamics $\mathcal{H}$.  For the purpose of this paper, we
assume that the hybrid trajectories exist, are as smooth as necessary, and are
discontinuous at countably infinite number of discrete switching times in $t$.
It is application specific as to whether the trajectories are left- or
right-side continuous.  A more careful treatment of existence,
differentiability, and other properties can be found in \cite{Bainov1993,
Rosa2022a}.

\subsection{Problem Statement}
\noindent \textit{Given}: An NLP with hybrid dynamics $\mathcal{H}$, $n$ grid
points, and $m$ phases of the form
\begin{equation}
\begin{array}{ll}
\underset{q_k, \dot{q}_k, \ddot{q}_k, u_k, f_k^i,
\iota_k^i, p}{\text{minimize}} \quad J(q_k, \dot{q}_k, \ddot{q}_k, u_k, f_k^i,
\iota_k^i, p) & \\
\begin{aligned}
\text{subject to} \;\; & M(q_k)\ddot{q}_k + b(q_k, \dot{q}_k) = u_k + B^i(q_k) f^i_k \\
& J^i(q_k) \ddot{q}_k + \dot{J}^i(q_k) \dot{q}_k + \phi^i(q_k, \dot{q}_k, t_k) = 0 \\
& M(q_{k_j})(\dot{q}_{k_i}^+ - \dot{q}_{k_j}^-) = \left(J^i(q_{k_j})\right)^T \iota_{k_i}^i\\
& J^i(q_{k_j})\dot{q}_{k_i}^+ = E^{ij} J^j(q_{k_j})\dot{q}_{k_j}^-\\
& c_L \leq c(q_k, \dot{q}_k, \ddot{q}_k, u_k, f_k^i, \iota_k^i, p) \leq c_U,
\end{aligned} &
\end{array}
\label{eqn:op}
\end{equation}
where $k, k_i, k_j$ are grid points ($0 \leq k, k_i, k_j \leq n$); $i$ is a
phase of the hybrid dynamics ($1 \leq i \leq m$); $q_k = q(t_k)$ is the value of
trajectory $q(t)$ at time $t = t_k$ (and similarly for $\dot{q}_k$,
$\ddot{q}_k$, $u_k$, $f_k^i$, and $\iota_k^i$); $M$, $b$, $B^i$, $J^i$,
$\dot{J}^i$, $\phi^i$, and $E^{ij}$ are as previously defined in \eqref{eqn:M}
and \eqref{eqn:I}; $p \in R^{n_p}$ is a vector of user-defined decision
variables---we implicitly assume all functions can depend on $p$, e.g., $M =
M(q; p)$; $c(\ldots) \in \R^{n_c}$ is a vector of user-defined constraint
functions; and $c_L \in \R^{n_c}$ and $c_U \in \R^{n_c}$ are the lower and upper
bounds of $c(\ldots)$.  Our notation and terminology mirrors that used in
\cite{Betts2010}.

\noindent \textit{Design}: A reproducible NLP that can be executed independently
of the library that generated it.

In the end, our approach to the problem is to implement common tasks that
traditionally rely on external libraries or non-generic implementations to
compute an optimal solution.  In this paper, we describe how to effectively bake
the 1) system dynamics, 2) ODE solver, and 3) polynomial reference trajectories
into the NLP model.  This introduces a new class of optimization software, where
the model is the library.


\section{Modeling Robot Constraints}
\label{sec:rc}
\subsection{A Unified Approach to Modeling Physical and Virtual Constraints}
\label{ssec:uni}
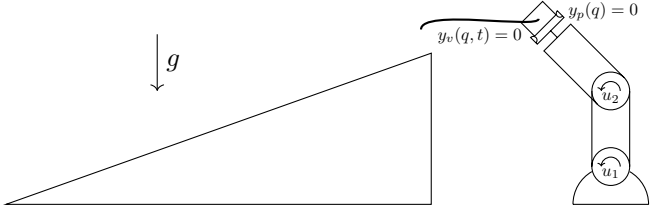
\begin{figure}[t]
\centering
\begin{tikzpicture}

  \draw (0.5, 0) arc [start angle=0, end angle=60, x radius=0.5cm, y radius=0.5cm]
    (-0.5, 0) arc [start angle=180, end angle=120, x radius=0.5cm, y radius=0.5cm]
    (-0.5, 0) -- (0.5, 0);

  \draw (0, 0.5) circle [radius=0.25cm] ++ (0.25, 0) -- 
    ++ (0, 1) ++ (-0.5, 0) -- ++(0, -1);

  \draw[-{>[length=0.05cm]}] (0.125, 0.5) 
    arc [start angle = 0, end angle=180, x radius = 0.125cm, y radius = 0.125cm] 
    node[pos=0.5, scale=0.65, below, yshift=-0.1cm] {$u_1$};

  \draw[rotate around={45:(0,1.5)}] (0, 1.5) circle [radius=0.25cm] ++ (0.25, 0) -- 
    ++ (0, 1) ++ (-0.5, 0) -- ++(0, -1);

  \draw[-{>[length=0.05cm]}] (0.125, 1.5) 
    arc [start angle = 0, end angle=180, x radius = 0.125cm, y radius = 0.125cm] 
    node[pos=0.5, scale=0.65, below, yshift=-0.1cm] {$u_2$};

  \draw[rotate around={45:(0,1.5)}] (0, 2.5) -- ++ (0, 0.125) -- 
    ++ (0.25, 0) -- + (0, 0.125) 
    arc [start angle=90, end angle=270, x radius=0.05cm, y radius=0.05cm]
    node[pos=0.5, right, scale=0.65, shift={(0.1cm,0.1cm)}] {$y_p(q) = 0$}
    ++ (-0.25, -0.025) -- ++ (-0.25, 0) -- + (0, 0.125)
    arc [start angle=90, end angle=-90, x radius=0.05cm, y radius=0.05cm]
    (-0.257, 2.5) -- (0.257, 2.5);

  \draw[rotate around={45:(0,1.5)}] (0, 2.5) +(-0.20, 0.125 + 0.1) rectangle
    +(0.20, 0.125 + 0.1 + 0.25);

  \draw[rotate around={45:(0,1.5)}, thick] (0, 2.5 + 0.125 + 0.1 + 0.25/2) .. controls
    +(-0.1, -0.1) and +(0.2, 0.2) .. +(-1.2, 1) 
    node[pos=0.5, below, scale=0.65] (virtual constraint) {$y_v(q, t) = 0$};

  \draw (-2.375, 2) -- (-2.375, 0) -- (-8, 0) -- cycle;

  \draw[->] (-6, 2.25) -- (-6, 1.5) node[pos=0.5, right] {$g$};

\end{tikzpicture}
\caption{An arm transports a rectangular object to a ramp.  The arm grasping the
object is modeled as a physical constraint $y_p(q) = 0$.  While the object is
grasped, the arm has to apply joint torques $u_1$ and $u_2$ in order to move the
object along a given path (the thick black curve).  The control law for tracking
the trajectory is computed through a virtual constraint $y_v(q, t) = 0$.}
\label{fig:arm}
\end{figure}
A unique contribution of our work is the generic treatment of physical and
virtual constraints.  Figure~\ref{fig:arm} highlights their differences.  In
this subsection and the next, we outline how our framework is capable of
modeling and stabilizing these types of constraints.  For simplicity in
notation, we drop the dependency on phase $i$ in this subsection and the next.

In our hybrid dynamics model $\mathcal{H}$, the term $J(q) \ddot{q} + \dot{J}
\dot{q} + \phi(q, \dot{q}, t) = 0$ of Equation~\ref{eqn:M} represents a class of
holonomic and nonholonomic constraints.  In the general case, these constraints
depend explicitly on state and time, are nonlinear in the configuration
variables, and linear in velocities and accelerations.

As an example, scleronomic constraints $y(q) = 0$ results in $J(q) =
\pd{y}{q}(q)$ and $\phi(q, \dot{q}, t) = 0$ for all time $t$.  The constraint
$y(q) = 0$ has been used to model contact constraints in manipulation and
locomotion problems.  Furthermore, contact constraints are examples of physical
constraints as the physical contact of two bodies will give rise to real-world
contact forces that resist interpenetration of the bodies in contact.  

For rheonomic constraints of the form $y(q, t) = h(q) + h_d(t)$, we have $J(q) =
\pd{h}{q}(q)$ and $\phi(q, \dot{q}, t) = \pd{^2h_d}{t^2}\left(t\right)$ for all
$t$.  A common application of this type of rheonomic constraint is in the design
of control laws of biped robots that enforce a set of virtual constraints
through feedback control \cite{Westervelt2007}.  The constraints are virtual
as they only exist while the robot's actuators are actively outputting the
computed control signal.

While physical and virtual constraints are conceptually different,  there is
computationally very little need to distinguish between the two with the right
level of abstraction.  For a generic approach to the two types of constraints,
let 
\begin{equation}
\begin{gathered}
J(q) = \begin{bmatrix} J_p^T(q), & J_v^T(q) \end{bmatrix}^T \in \R^{(n_p + n_v) \times n_q}, \\
B(q) = \begin{bmatrix} J_p^T(q), & B_v(q) \end{bmatrix} \in \R^{n_q \times (n_p + n_u)},
\end{gathered}
\label{eqn:JB}
\end{equation}
where $n_p$ and $n_v$ are the number of physical and virtual constraints,
respectively; $J_p$ and $J_v$ are constraint Jacobians; $n_u$ are the number of
control inputs, and $B_v$ is a transmission matrix mapping constraint forces to
generalized forces.  For $n_v$ independent virtual constraints, there needs to
be $n_u \geq n_v$ control inputs.

It is straightforward to verify that if $J(q)$ and $B(q)$ of
Equation~\ref{eqn:JB} have maximal rank at each point in time throughout a
trajectory of $\mathcal{H}$, then we can plug Equation~\ref{eqn:JB} into
Equation~\ref{eqn:M} and simultaneously solve for a set of physical and virtual
constraints.  However, nonzero output error (e.g., $y(q,t) \neq 0$) is a common
issue when trying to enforce constraints at the acceleration level.  In the next
subsection, we outline how output error can be mitigated in a unified manner.

\subsection{A Unified Approach to Constraint Stabilization}
When solving for the dynamics of a constrained system (and once again ignoring
the phase dependency $i$), it is often desirable to have some form of constraint
stabilization to avoid the accumulation of nonzero output error in the
constraint equations.  Without loss of generality, assume a holonomic constraint
of the form $y(q, t) = \left[ \begin{smallmatrix} y_p(q, t) \\ y_v(q, t)
\end{smallmatrix} \right] = 0$, where $y_p(q, t) \in \R^{n_p}$ and $y_v(q, t)
\in \R^{n_v}$ are physical and virtual holonomic constraints, respectively.
Then, the linear control law, 
\begin{equation}
\begin{aligned}
v(q, \dot{q}, t) &= \begin{bmatrix} v_p^T(q, \dot{q}, t), 
                     & v_v^T(q, \dot{q}, t) \end{bmatrix}^T \in \R^{n_p + n_v} \\
                 &= K_P y(q, t) + K_D \dot{y}(q, t)
\end{aligned}
\label{eqn:v}
\end{equation}
drives small errors in the constraint equations $y(q,t) = 0$ and $\dot{y}(q,t) =
0$ to zero for appropriately chosen gain matrices $K_P \in \R^{(n_p+n_v) \times
(n_p+n_v)}$ and $K_D \in \R^{(n_p+n_v) \times (n_p+n_v)}$.  A similar statement
holds if the constraints are not rheonomic or holonomic (e.g.,
\cite{Griffin2015}).

For physical constraints ($n_p > 0$), the control law $v_p(q, \dot{q}, t)$ is
Baumgarte's stabilization method \cite{Bauchau2007}.  For virtual constraints
($n_v > 0$), $v_v(q, \dot{q}, t)$ is the stabilizing controller of an
input-output linearizing control law \cite{Khalil2002}.  In particular, for
biped robots, it can be shown that $v_v(q, \dot{q}, t)$ is the control law for
driving a solution onto the zero manifold in the hybrid zero dynamics framework
\cite{Westervelt2007}.

In summary, when Equations~\ref{eqn:JB} and \ref{eqn:v} are substituted into
Equation~\ref{eqn:M}, we have during a phase
\begin{equation}
\begin{gathered}
M(q)\ddot{q} + b(q, \dot{q}) = u + J_p^T(q) f_p + B_v(q) f_v \\
J_p(q) \ddot{q} + \dot{J}_p(q) \dot{q} + \phi_p(q, \dot{q}, t) = 0, \\
J_v(q) \ddot{q} + \dot{J}_v(q) \dot{q} + \phi_v(q, \dot{q}, t) = 0,
\end{gathered}
\label{eqn:JpJvBv}
\end{equation}
where $f_p$ is a vector of physical forces and $f_v$ is a vector of input-output
linearizing feedback control laws for enforcing the virtual constraints such
that $f$ of Equation~\ref{eqn:M} is $f = [f_p^T, f_v^T]^T$.  If we assume
constraints of the form $y(q, t) = 0$, then $\phi_p(q, \dot{q}, t) =
\pd{^2y_p}{t^2}\left(q,t\right) + 2 \pd{^2y_p}{t \partial q}\left(q,t\right) +
v_p(q, \dot{q}, t)$ and similarly for $\phi_v(q, \dot{q}, t)$.

In Section~\ref{ssec:tja}, we show how to compute $J_p(q)$, $J_v(q)$,
$\dot{J}_p(q) \dot{q}$, $\dot{J}_v(q) \dot{q}$, and $B_v(q)$ as a set of
constraints in a nonlinear programming problem.  Quantities that explicitly
depend on time, such as $\pd{^2y}{t^2}\left(q,t\right)$, have to be computed by
the user or other modules in the Amplify library.  We give an example in
Section~\ref{sec:ex} that implements virtual rheonomic constraints of the form
$y(q, t) = h(q) - h_d(t) = 0$ for a hybrid zero dynamics optimization problem.
The example uses the algorithm in Section~\ref{ssec:tja} and Amplify's
B\'{e}zier module (Section~\ref{sec:bez}).


\section{Describing Constrained Mechanical Systems with Spatial Vectors}
\label{sec:sv}
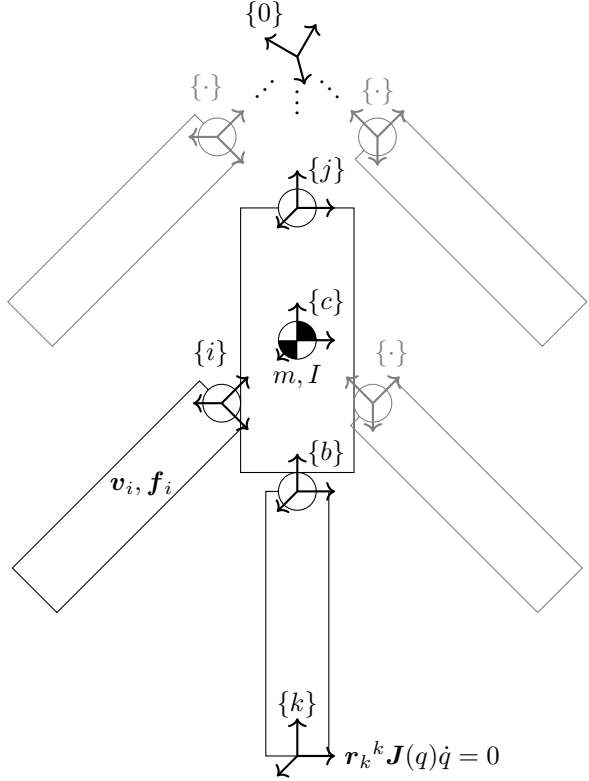
\begin{figure}[t]
\centering
\begin{tikzpicture}
  \def \r{0.25cm}
  \def \w{0.5cm}
  \def \h{3.5cm}

  \def \rigidbody[#1]{circle [radius=\r] ++ (\r, 0) 
    -- ++(#1, 0) -- ++(0, -\h) -- ++(-2*\r - 2*#1, 0) -- ++ (0, \h) -- ++ (#1, 0) 
    ++ (2 * \r, 0) -- cycle}

  \def \axes#1#2#3{%
    \draw[thick, ->] (#1) -- +(0.5, 0, 0);%
    \draw[thick, ->] (#1) -- +(0, 0.5, 0) node[#2] {$\{#3\}$};%
    \draw[thick, ->] (#1) -- +(0, 0, 0.7)%
  }

  \coordinate (r) at (0, 3 / 2 * \h + 2 * \r);
  \coordinate (j) at (0, \h + \r);
  \coordinate (b) at (0, 0);
  \coordinate (k) at ($(b) + (0, -\h)$);
  \coordinate (i) at (-2 * \r - \w, \h / 3);
  \coordinate (child) at (2 * \r + \w, \h / 3);
  \coordinate (com) at ($(j) + (0, -\h /2)$);

  \coordinate (main branch) at ($(r) + (-90:1.5cm)$);
  \coordinate (left branch) at ($(r) + (-135:1.5cm)$);
  \coordinate (right branch) at ($(r) + (-45:1.5cm)$);

  \begin{scope}[rotate around={60:(r)}]
    \axes{r}{above}{0};
  \end{scope}
  \node[anchor=center] at ($(r) + (-90:0.5cm)$) {$\vdots$};
  \node[anchor=center, rotate=45] at ($(r) + (-45:0.5cm)$) {$\vdots$};
  \node[anchor=center, rotate=-45] at ($(r) + (-135:0.5cm)$) {$\vdots$};

  \draw[gray] (left branch) [rotate around={-45:(left branch)}] \rigidbody[\w / 3];
  \begin{scope}[gray, rotate around={-45:(left branch)}]
    \axes{left branch}{above, xshift=-0.5cm}{\cdot};
  \end{scope}

  \draw[gray] (right branch) [rotate around={45:(right branch)}] \rigidbody[\w / 3];
  \begin{scope}[gray, rotate around={45:(right branch)}]
    \axes{right branch}{right, yshift=0.25cm}{\cdot};
  \end{scope}

  \draw (j) \rigidbody[\w];
  \axes{j}{right}{j};

  \fill (com) -- ++(\r, 0) 
    arc [x radius=\r, y radius=\r, start angle=0, end angle=90]
    -- ++(0, -2 * \r) 
    arc [x radius=\r, y radius=\r, start angle=270, end angle=180] -- cycle;

  \draw (com) circle [radius = \r] node[below, yshift=-0.2cm] {$m, I$};
  \axes{com}{right}{c};

  \draw (b) \rigidbody[\w / 3];
  \axes{b}{right}{b};
  \axes{k}{pos=0.25, above,yshift=0.25cm}{k};
  \node[right,xshift=0.5cm] at (k) {$\spatr_k \spat[k]{J}(q) \dot{q} = 0$};

  \begin{scope}[rotate around={-45:(i)}]
    \draw (i) \rigidbody[\w / 3];
    \axes{i}{above,xshift=-0.5cm}{i};
    \node at ($(i) + (0, -1.5cm)$) {$\spatv_i, \spatf_i$};
  \end{scope}

  \begin{scope}[gray, rotate around={45:(child)}]
    \draw (child) \rigidbody[\w / 3];
    \axes{child}{above,xshift=0.6cm}{\cdot};
  \end{scope}

\end{tikzpicture}
\caption{An articulated rigid-body system.}
\label{fig:kt}
\end{figure}

We assume Equations \eqref{eqn:M}--\eqref{eqn:I} are derived from an articulated
rigid-body system of links and joints.  Such a system can be represented as a
rooted kinematic tree with links as nodes and joints as edges.  Traversal of the
tree naturally leads to a recursive formulation of the equations of motion
\cite{Featherstone2010, Lynch2023}.

We compute the equations of motion using the 
\begin{itemize}
\item Recursive Newton-Euler Algorithm (RNEA), which computes $b(q, \dot{q})$;

\item Composite Rigid Body Algorithm (CRBA), which computes $M(q)$; and 

\item Task Jacobian Algorithm (TJA), which computes $J^i(q)$, $B^i(q)$,
$\dot{J}_p(q) \dot{q}$, and $\dot{J}_v(q) \dot{q}$.

\end{itemize}
With our extensions, the TJA also computes position and velocity constraints and
the constraint stabilization control law of Equation~\ref{eqn:v}.  We implement
the three algorithms using the spatial vector algebra of
\cite{Featherstone2008}.  These algorithms are typically implemented using an
imperative programming style (e.g., \cite{Featherstone2010b}).  In
Section~\ref{sec:rbd}, we formulate the algorithms as a series of constraints.
In the remainder of this section, we present a brief overview of spatial vectors
for use as a reference throughout the paper.

\subsection{Spatial Velocities, Forces, and Frames}
\label{ssec:spatv}
The motion of a single link can be represented with an angular velocity vector
$\omega = [\omega_x, \omega_y, \omega_z]^T \in \R^3$ and a linear velocity
vector $v = [v_x, v_y, v_z]^T \in \R^3$ about the coordinate $x$-$y$-$z$ axes
of an inertial frame fixed in space \cite{Featherstone2008, Lynch2023}.  A
similar statement exists for torques $\tau \in \R^3$ and forces $f \in \R^3$
acting on the rigid body (see Figure~\ref{fig:kt}).

A spatial vector algebra treats angular and linear motions and forces as 6D
vectors in dual spaces.  As coordinate vectors in Cartesian space, these vectors
reside in $\R^6$.  Example 6D motion vectors are spatial velocities $\spatv =
[\omega^T, v^T]^T$ and accelerations $\spata = \spatdot{}{v}{}$.  Example 6D
force vectors are spatial forces $\spatf = [\tau^T, f^T]^T$, impulses
$\spat{\iota}$, and momenta $\spatI \spatv$; the quantity $\spatI \in \R^{6
\times 6}$ is the spatial inertia of a rigid body (see
Section~\ref{ssec:spatI}).

In terms of notation, spatial quantities are in bold text, a subscript ${i \in
\mathbb{N}}$ on a coordinate vector denotes the spatial quantity of the
$i^\text{th}$ link, e.g., $\spatv_i$ is the spatial velocity of the
$i^\text{th}$ link of the robot.  A superscript is the frame the quantity is
represented in, e.g., $\spat[j]{v}[i]$ is the velocity of the $i^\text{th}$ link
represented in the $j^\text{th}$ frame ($j \in \mathbb{N}$; no superscript
implies the body frame, e.g., $\spatv_i = \spat[i]{v}[i]$).  Frames are
represented with curly brace notation, e.g., the $i^\text{th}$ frame is frame
$\{i\}$ \cite{Lynch2023}.

Finally, we follow the convention in \cite{Featherstone2008}, where link $i$ and
joint $i$ share the same frame $\{i\}$.  The root's frame, which is always
$\{0\}$ in Amplify, is the world frame of the system.  Other frames in our work
are the center-of-mass (COM) frames $\{c\}$ and constraint frames $\{k\}$.  In
practice, COM and constraint frames are always defined relative to a link frame.
For constraints, we use link-frame $\{b\}$ as a mnemonic that we are working
with constraints.  In particular, frames $\{k\}$ and $\{b\}$ should be
considered as a pair $(k, b)$.  Furthermore, link and frame labels (e.g., $i, j,
k, b \in \mathbb{N}$) also serve as indices into multidimensional lists of
spatial quantities, which is common practice in rigid body dynamics algorithms
\cite{Featherstone2010}.

\subsection{Spatial Transforms and Cross Products}
\label{ssec:spatX}
For motion vectors, the spatial transform $\spat[i]{X}[j] \in \R^{6 \times 6}$,
a coordinate transformation matrix, maps a vector in frame $\{j\}$ coordinates
into frame $\{i\}$ coordinates, e.g., $\spatv_i = \spat[i]{X}[j] \spatv_j$.
Force vectors transform according to ${\spat[i]{X}[j][*] = \spat[i]{X}[j][-T] =
\spat[j]{X}[i][T]}$, e.g., $\spatf_i = \spat[i]{X}[j][*] \spatf_j$.

We represent spatial transforms using axis-angle notation $(\hat{a}, \theta) \in
\R^7$ \cite{Kim2023}.  As a product, $\hat{a} \theta$ is a twist with normalized
vector $\hat{a} = [\hat{\omega}^T, \hat{v}^T]^T \in \R^6$ and magnitude $\theta
\in \R$.  The spatial transform $^i\spatX_j$ is then
\begin{equation}
^i\spatX_j = \begin{bmatrix} E, & 0 \\ p \times E, & E  \end{bmatrix},
\label{eqn:X}
\end{equation}
where $E \in SO(3)$ is a rotation matrix and $p \in \R^3$ is a position vector
pointing from $\{i\}$ to $\{j\}$ such that
\begin{equation}
\begin{aligned}
E &= I_3 - \sin(\theta) {\hat{\omega} \times} + (1 - \cos(\theta)) {\hat{\omega} \times} {\hat{\omega} \times} \\ 
p &= (I_3 \theta - (1 - \cos(\theta)) {\hat{\omega} \times} + (\theta -
      \sin(\theta)) {\hat{\omega} \times} {\hat{\omega} \times}) \hat{v}.
\end{aligned}
\label{eqn:Ep}
\end{equation}

\begin{myrem}
Our equations of Rodrigues' formula for $E$ and the position vector $p$ are not
typos.  The definition of a positive rotation is different than how it is
traditionally defined (e.g., compare \cite{Featherstone2008} and
\cite{Lynch2023}).  We define positive rotations to be consistent with
\cite{Featherstone2008}.
\end{myrem}

The 3D cross products in Equations~\ref{eqn:X} and \ref{eqn:Ep} can be written
as a skew-symmetric matrix:
\begin{equation}
{\omega \times} = \begin{bmatrix} 0, & -\omega_z, & \omega_y \\
                              \omega_z, & 0, & -\omega_x \\
                              -\omega_y, & \omega_x, & 0  \end{bmatrix}, \quad
\omega = \begin{bmatrix} \omega_x \\ \omega_y \\ \omega_z \end{bmatrix}.
\end{equation}

There also exists two 6D cross products for spatial terms.  The cross product of
a motion vector with another motion vector is:
\begin{equation}
\spatv \times \spat{m} = \begin{bmatrix}
                        {\omega \times} & 0 \\
                        {v \times} & {\omega \times}
                         \end{bmatrix} \spat{m}, \quad  
\spatv = \begin{bmatrix} \omega \\ v \end{bmatrix}.
\end{equation}
For forces, we have ${\spatv \times^*} \spatf$, where ${\spatv \times^*} =
{-\spatv \times}^T$.

The derivative of a spatial transform is 
\begin{equation}
\spatdot{i}{X}{j} = {\left( \spat[i]{X}[j] \spatv_j - \spatv_i \right) \times} \spat[i]{X}[j],
\end{equation}
where $\spatv_j$ and $\spatv_i$ are the spatial velocities of $\{j\}$ and
$\{i\}$, respectively.  Further details regarding 3D and 6D cross products and
their use in computing derivatives can be found in \cite{Featherstone2008}.

Finally, we can extract the relative orientation and position of two frames
given their spatial transform.  Let $\spatX = \left[\begin{smallmatrix} X_1 & 0
\\ X_2 & X_1 \end{smallmatrix}\right]$, where $X_1, X_2 \in \R^{3 \times 3}$ are
submatrices of Equation~\ref{eqn:X}, then
\begin{equation}
\begin{aligned}
\ori(\spatX) &= \begin{bmatrix}
          \arctan(-[X_1]_{23}, [X_1]_{33}), \\
          \arcsin(-[X_1]_{13}), \\
          \arctan(-[X_1]_{12}, [X_1]_{11})
          \end{bmatrix} \\
\pos(\spatX) &= \begin{bmatrix} [-X_1^T X_2]_{32}, \\ [-X_1^T X_2]_{21}, \\
[-X_1^T X_2]_{13}  \end{bmatrix},
\end{aligned}
\label{eqn:X}
\end{equation}
where $\ori(\spatX) \in \R^3$ returns $x$-$y$-$z$ Euler angles and $\pos(\spatX)
\in \R^3$ returns a point $p = (x, y, z)$ in Cartesian coordinates, and the
notation $[X]_{rc} \in \R$ represents the value in row $r$ and column $c$ of the
matrix $X$.  If the input quantity is a vector, then a single index is used to
represent the value to extract.

\begin{myrem}
We opted to decompose $\spatX$ into Euler angles and a position vector as
opposed to an axis-angle pair because of its straightforward implementation in a
modeling language.  The representation suffers from singularities whenever the
Euler angle about the $y$-axis approaches $\pm \frac{\pi}{2}$.
\end{myrem}

\subsection{Spatial Inertias}
\label{ssec:spatI}
The spatial inertia of a link about its center of mass $\spatI_c$ is
\begin{equation}
\spatI_c = \begin{bmatrix} I, & 0 \\ 0, & m I_3 \end{bmatrix}, \quad
I = \begin{bmatrix} I_{xx}, & I_{xy}, & I_{xz} \\ 
                    I_{xy}, & I_{yy}, & I_{yz} \\
                    I_{xz}, & I_{yz}, & I_{zz}
    \end{bmatrix}
\label{eqn:spatI}
\end{equation}
where $m \in \R$ is the mass of the body and scalars $I_{xx}$, $I_{yy}$,
$I_{zz}$, $I_{xy}$, $I_{xz}$, and $I_{yz}$ form the moment of inertia tensor.
Spatial inertias map to different frames according to $\spat[i]{I}[j] =
\spat[i]{X}[j][*] \spat{I}[j] \spat[j]{X}[i] = \spat[j]{X}[i][T]
\spat{I}[j] \spat[j]{X}[i]$.

\subsection{Spatial Jacobians}
\label{ssec:spatr}
For a constrained mechanical system in phase $i$, we represent the $r^\text{th}$
constraint that restricts the motion of a body-fixed point coincident with frame
$\{k\}$ in a link $b$ as 
\begin{equation}
\spatr^{r}_k \spat[k]{J}(q) \dot{q}  = 0 \in \R,
\label{eqn:rv}
\end{equation}
where $\spatr^r_k \in \R^6$ is a constraint vector and $\spat[k]{J}(q) \in \R^{6
\times n_q}$ is the spatial Jacobian.  The spatial Jacobian maps generalized
velocities $\dot{q}$ to the spatial velocity at frame $\{k\}$ such that
\begin{equation}
\begin{aligned}
\spat[k]{J}(q) &=
\begin{bmatrix}
  \spat[k]{s}[1](q), & \ldots, & \spat[k]{s}[j](q), & \ldots, & \spat[k]{s}[n_q](q) 
\end{bmatrix} \\
\spat[k]{s}[j](q) &=
\begin{cases}
  \spat[k]{X}[b] \spat[b]{X}[j](q) \spats_j, & j \in \osi{A}[b] \\
  0, & \text{otherwise}
\end{cases}.
\end{aligned}
\label{eqn:spatJ}
\end{equation}
Together, these equations state that the generalized velocities that contribute
to the spatial velocity at frame $\{k\}$ in link $b$ are on the path from frame
$\{b\}$ to the root of the tree (i.e., $j \in \osi{A}[b]$).  The relative joint
velocities at unit speed in frame $\{k\}$ coordinates $\spat[k]{s}[j](q)$ form
the nonzero columns of $\spat[k]{J}(q)$.  We avoid the zero entries in
Equation~\ref{eqn:spatJ} by leveraging the results in \cite{Featherstone2010b},
which preserves the branch-induced sparsity pattern of the kinematic tree when
computing the constraint Jacobian $J^i(q)$.  

When it comes to computing the constraint Jacobian $J^i(q)$, the spatial
coordinates we compute the Jacobian in matters.  Equation~\ref{eqn:spatJ}
computes spatial constraints in a body-fixed coordinate system $\{b\}$, which is
a common reference frame for certain optimization problems (e.g.,
\cite{Balkcom2000}).  However, as pointed out in \cite{Featherstone2010}, the
typical intent of $J^i(q)\dot{q} = 0$ in robotics is to constrain the motion at
a particular point $p$ relative to the world frame.  In this case, the spatial
Jacobian becomes $\spat[\bar{k}]{J}(q) = \spat[0]{E}[k](q) \spat[k]{J}(q)$,
where $\{\bar{k}\}$ is the frame located at the same point $p$ but with axes
aligned with the world frame $\{0\}$ axes.  In other words, the equivalent
constraint of Equation~\ref{eqn:rv} in a world-aligned frame is computed in a
frame $\{\bar{k}\}$.  We compute both types of constraints at $\{k\}$ and
$\{\bar{k}\}$ in Amplify and describe our implementation in
Section~\ref{sec:rbd}.


\section{The Grid Layout}
\label{sec:grid}
\begin{table}[t]
\begin{center}
\scriptsize
\begin{tabular}{c|c|c}
\hline
Sets & Definition & Description \\
\hline
$\us{S}$ & specified as input, $\R \setminus \{0\}$ & subphases \\
$\us{P}$ & $\left\{ \lfloor j \rfloor: j \in \us{S} \right\}$ & phases \\
$\usi{S}$ & $\left\{ j \in \us{S}: \lfloor j \rfloor = i
\right\}$ & subphases grouped by phase \\
$(\gosi{G}, \leq)$ & $\bigcup_{j \in \usi{S}} [a_j, b_j] \subset \mathbb{N} $ & grid points grouped by phase \\
$(\gosi{G}[0], \leq)$ & $\left[0, n_{G_\text{max}}\right] \cap \bigcup_{i \in
\us{P}} \usi{G}$ & all grid points 
\end{tabular}
\end{center}
\caption{Definition of grid sets.  Ordered sets are denoted as a pair (see
Table~\ref{tab:kt} for explanation).  Otherwise, the set has no ordering.}
\label{tab:grid}
\end{table}
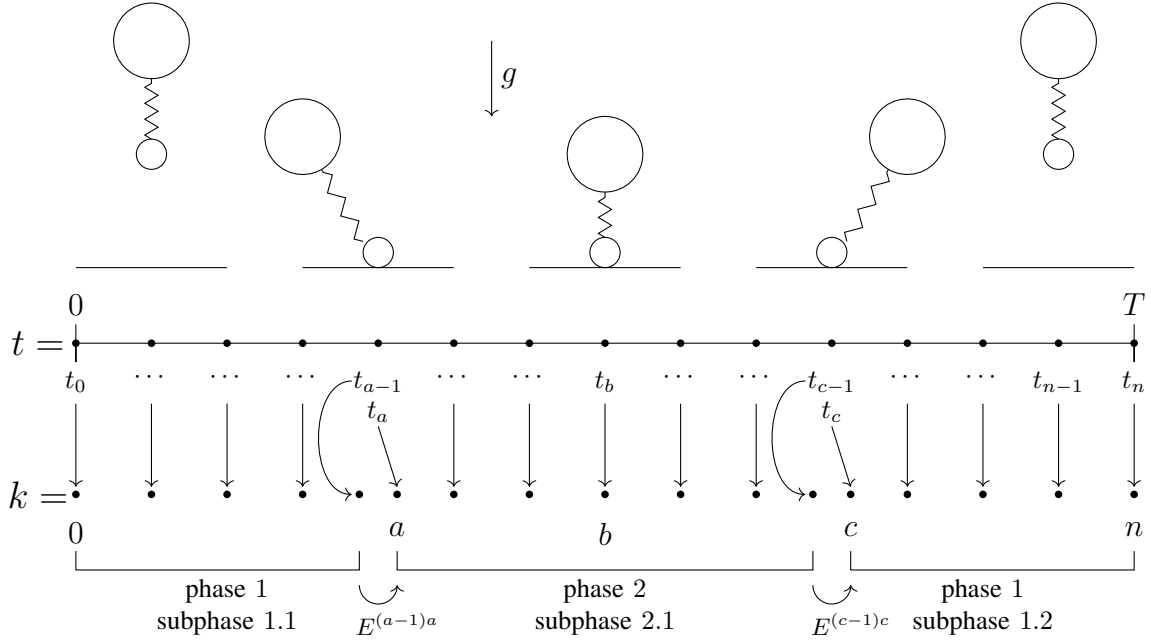
\begin{figure*}[t]
\centering
\begin{tikzpicture}[decoration=zigzag]
  \def \r{0.2}
  \def \R{0.5}
  \def \d{0.25}
  \def \p{0.05}
  \def \e{0.25}

  \draw (-7, 0) -- (-5, 0);
  \draw (-4, 0) -- (-2, 0);
  \draw (-1, 0) -- (1, 0);
  \draw (2, 0) -- (4, 0);
  \draw (5, 0) -- (7, 0);

  \draw (-6, 1.5) circle [radius=\r];
  \draw (-3, \r) circle [radius=\r];
  \draw (0, \r) circle [radius=\r];
  \draw (3, \r) circle [radius=\r];
  \draw (6, 1.5) circle [radius=\r];

  \draw (-6, 3) circle [radius=\R];
  \draw[rotate around={30:(-3,0)}] (-3, 2) circle [radius=\R];
  \draw (0, 1.5) circle [radius=\R];
  \draw[rotate around={-30:(3,0)}] (3, 2) circle [radius=\R];
  \draw (6, 3) circle [radius=\R];

  \draw[decorate, segment length=6] (-6, 1.5 + \r) -- (-6, 3 - \R);
  \draw[decorate, rotate around={30:(-3,0)}] (-3, 2 * \r) -- (-3, 2 - \R);
  \draw[decorate, segment length=6] (0, 2*\r) -- (0, 1.5 - \R);
  \draw[decorate, rotate around={-30:(3,0)}] (3, 2 * \r) -- (3, 2 - \R);
  \draw[decorate, segment length=6] (6, 1.5 + \r) -- (6, 3 - \R);

  \node[font=\Large] at (-7.5, -1) {$t =$};
  \node[above=0.25, font=\large] at (-7, -1) {$0$};
  \node[above=0.25, font=\large] at (7, -1) {$T$};
  \draw (-7, -1) -- +(0, -\d) -- +(0, \d) +(0, 0) 
    -- (7, -1) -- +(0, -\d) -- +(0, \d);

  \foreach \i in {-7,...,7} {
    \fill (\i,-1) circle [radius=\p];
  }

  \foreach \i in {-6,...,-4,-2,-1,1,2,4,5} {
    \node[below=0.25] at (\i, -1) {$\cdots$};
  }

  \node[below=0.25] at (-7, -1) {$t_0$};
  \node[below=0.25] at (-3, -1) {\shortstack{$t_{a-1}$\\$t_a$}};
  \node[below=0.25] at (0, -1) {$t_b$};
  \node[below=0.25] at (3, -1) {\shortstack{$t_{c-1}$\\$t_c$}};
  \node[below=0.25] at (6, -1) {$t_{n-1}$};
  \node[below=0.25] at (7, -1) {$t_n$};

  \node[font=\Large] at (-7.5, -3) {$k = $};
  \node[below=0.25, font=\large] at (-7, -3) {$0$};
  \node[below=0.25, font=\large] at (-3+\e, -3) {$a$};
  \node[below=0.25, font=\large] at (0, -3) {$b$};
  \node[below=0.25, font=\large] at (3+\e, -3) {$c$};
  \node[below=0.25, font=\large] at (7, -3) {$n$};

  \foreach \i in {-7,...,7} {
    \pgfmathparse{abs(\i)==3 ? 1 : 0};
    \ifnum\pgfmathresult=1
      \fill (\i-\e,-3) circle [radius=\p];
      \draw[->,thin] (\i-0.35,-1.5) to[out=180,in=180] (\i-0.35,-3);
      \fill (\i+\e,-3) circle [radius=\p];
      \draw[->,thin] (\i,-2.1) -- (\i+0.25,-2.9);
    \else
      \draw[->,thin] (\i,-1.8) -- (\i,-2.9);
      \fill (\i,-3) circle [radius=\p];
    \fi
  }

  \node[below, align=center, font=\normalsize] at (-5, -4) {phase $1$\\subphase $1.1$};
  \draw (-7, -4) -- +(0, \d) +(0, 0) 
    -- (-3-\e, -4) -- +(0, \d);

  \node[below, align=center, font=\normalsize] at (0, -4) {phase $2$\\subphase $2.1$};
  \draw (-3+\e, -4) -- +(0, \d) +(0, 0) 
    -- (3-\e, -4) -- +(0, \d);

  \node[below, align=center, font=\normalsize] at (5, -4) {phase $1$\\subphase $1.2$};
  \draw (3+\e, -4) -- +(0, \d) +(0, 0) 
    -- (7, -4) -- +(0, \d);

  \draw[->,thin] (-3-\e, -4.2) arc [start angle=180,end angle = 360,radius=\e]
    node[below=0.25,font=\small] {$E^{(a-1)a}$};
  \draw[->,thin] (3-\e, -4.2) arc [start angle=180,end angle = 360,radius=\e]
    node[below=0.25,font=\small] {$E^{(c-1)c}$};

  \draw[->,font=\large] (-1.5, 3) -- (-1.5, 2) node[pos=0.5, right] {$g$};

\end{tikzpicture}
\caption{An example grid layout for a single-legged hopping robot with $t \in
[0, T]$ corresponding to continuous time, $t_0$,$\ldots$,$t_n$ discretized time,
and $k \in [0,\ldots,n]$ the grid points.  The discretized time points are shown
mapped to their corresponding grid points.  Discrete times $t_{a-1}$ and $t_a$
are switching times such that $t_{a-1}=t_a$ and similarly for $t_{c-1}$ and
$t_c$.  Overall, the hybrid dynamics is modeled with two phases, 1 and 2, and
three subphases, 1.1, 2.1, and 1.2.  Phase 1 corresponds to the dynamics of the
robot in free flight.  Phase 2 represents the robot pivoting about its foot on
the ground.  Subphases 1.1 and 1.2 share the same free-flight dynamics, but one
subphase corresponds to flight prior to touchdown and the other after takeoff.
The impulse-momentum equations maps the state in one subphase to a state in a
different subphase through the coefficient matrices $E^{(a-1)a}$ and
$E^{(c-1)c}$, respectively.  Additional constraints can be added pointwise.  For
example, at grid point $b$, the ground reaction force can be constrained to be
less than a max desired value.}
\label{fig:grid}
\end{figure*}
In our framework, we optimize trajectories over a set of grid points
$\gosi{G}[0]$ in an interval of $[0, n_\text{max}] \subset \R$.  Grid points do
not represent the evolution of time, but a set of constraints that a solver
needs to satisfy at the specific grid point.  A grid point can have any set of
shared or unique constraints applied to it relative to other grid points.

For example, instantaneous impact are often modeled as occurring at the same
point in time across distinct pre- and post-impact events.  As constraints in
Amplify, a pre-impact collision constraint is applied at a unique grid point in
$\gosi{G}[0]$ while the corresponding post-impact collision constraint is
applied at a different grid point (e.g., grid points $a-1$ and $a$ in
Figure~\ref{fig:grid}).  These grid points share a switching time constraint
that makes absolute time equal at these two points.  

For trajectory optimization problems, it is important to group grid points
together to form trajectories that satisfy the equations of motion of a hybrid
arc.  To apply a common set of equations of motion, grid points are partitioned
into user-defined phases and subphases.
\begin{mydef}
A phase $i \in \us{P}$ represents a union of subphases that are subject to the
same set of physical and virtual constraints across consecutive grid points in
$\gosi{G}[0]$.  While physical and virtual constraints can change across phases,
all grid points are subject to the same unconstrained equations of motion as
described in Section~\ref{sec:H}.
\end{mydef}

\begin{mydef}
A \emph{subphase} is a single consecutive set of grid points $[a_j, b_j] \subset
\gosi{G}[0]$ ($a_j \leq b_j$) with labels ${j \in \us{S} \subset \R}$.  A
subphase is used to apply a set of additional constraints on the interval $[a_j,
b_j]$ beyond the equations of motion imposed by the phase dynamics.  For
example, the desired foot placement of a biped robot can be different for each
subphase while the overall step-to-step dynamics are the same (e.g.,
\cite{Wang2024}).
\end{mydef}

In terms of implementation, phases are derived from subphases through a mapping
of the floor function. For example, subphase $j = 1.1$, $1.2$, or $1.971$ all
map to phase $i = 1$.  This leads to a partition of grid points by phase, where
$\gosi{G} = \bigcup_{j \in \usi{S}} [a_j, b_j]$ is an indexed set of grid points
and $\usi{S}$ is an indexed set of subphases associated with phase $i$.
Table~\ref{tab:grid} gives the definition of these sets.

Figure~\ref{fig:grid} provides a visual overview of an example grid.  In this
example, a single-legged hopping robot is modeled with two phases and three
subphases.  The two phases correspond to the dynamics of the robot pivoting
about its foot on the ground and free flight.  Two of the subphases share the
same free-flight dynamics.  The first subphase includes a landing event and the
latter a takeoff event.  The third subphase is governed by the the
foot-on-the-ground dynamics.  The impulse-momentum equations (IME) connect the
state in one subphase to a state in a different subphase.  For example, the IME
adds constraints at the beginning of the landing phase that maps the velocity at
the end of the free-flight phase to the initial velocity during the
foot-on-the-ground phase.

\subsection{Modularity}
Given the lack of specific language constructs for modular design in AMPL, we
rely on convention to define how a module should be written and used.  This
approach to modularity and encapsulation is similar to languages such as
Mathematica and python, where users and developers enter into a tacit agreement
that certain variables and constraints should not be modified for the overall
correct operation of the module.

In general, modules are blocks of NLP code.  They specify constraints, parse
their module-specific inputs, and define their own private parameters and
decision variables that other modules should not touch.  They can also define
shared decision variables that other modules can assign values to through their
own module-specific constraints.  For example, the generalized accelerations
$\ddot{q}(t)$ and constraint forces $f^i(t)$ of Equation~\ref{eqn:M} are shared
decision variable that any module can constrain at any point in time $t$ along
the variable's trajectory.

Analogous to memory allocation in a computer program, modules specify the grid
points over which their variables and constraints are applied at.  The module
chooses the grid points based on user input or pre-defined sets of grid points
defined in Amplify (e.g., $\gosi{G}[0]$).  To simplify the module's design, any
decision variable specified at an integer grid point is considered part of
Amplify's shared workspace such that the variables can be constrained in a
different module.  This has the added advantage that when a seed value converges
to an optimal solution, decision variables with values at integer grid points
are considered part of the final trajectory.

Grid points that occur at non-integer values are meant to be used for
intermediate computations that may or may not be useful to extract as part of
the final solution.  This all depends on the intention of the module that
created the grid point.  For example, if the trajectory is meant to be
interpolated, then the intermediate computations of the ODE module would be
useful to extract from an optimal solution (see \cite{Kelly2017}).

Finally, we have a few mechanims for controlling if a module is active in a
particular instance of an NLP.  A pre-requisite for inclusion is a nonempty set
of grid points in $\gosi{G}[0]$ that the module's variables and constraints are
assigned to.  However, the primary mechanism is the enumeration of all variables
and constraints that will be sent to a solver through AMPL's \texttt{problem}
keyword.  If a decision variable or constraint is omitted, the decision variable
is fixed or the constraint is dropped.  While a burden to do by hand, it has the
advantage of explicitly documenting the specific NLP the user intends to solve.

Overall, these design choices permit modules to be swapped out with other
modules that have different implementations, but the same input/output
structure.  They can also be combined with other modules with similar design
rules to create a complete NLP from smaller building blocks.  Longer term, we
envision an ecosystem where modular blocks of NLP code are part of a NLP code
repository and package management system similar to the node package manager
\verb+npm+.  This is in contrast to the monolithic libraries developed today,
where the NLP statements are not treated as first-class entities.


\section{Rigid-Body Dynamics Algorithms as Constraints}
\label{sec:rbd}
In this section, we describe how the data structures (e.g., the kinematic tree)
and algorithms of classic rigid-body dynamics algorithms (i.e., the
Recursive-Newton Euler, Composite Rigid Body, and Task Jacobian algorithms)
can be implemented as constraints in modeling languages that permit
conditionals and recursive definitions of their decision variables and
parameters.

\subsection{Representing the Connectivity of the Kinematic Tree}
\label{ssec:kt}
\begin{table}[t]
\begin{center}
\scriptsize
\begin{tabular}{c|c|c}
\hline
Ordered Set & Definition & Description \\ \hline
$(\mathrm{L}_0, \leq)$ & $\left( 0, 1, \ldots, n_q \right)$ & link from root to leaves \\
$(\os{L}_1, \leq)$ & $\left( 1, 2, \ldots, n_q \right)$ & links excluding root \\
$(\os{L}_{-1}, \geq)$ & $\left( n_q, n_q - 1, \ldots, 0 \right)$ & links from leaves to root \\ 
\hline \\[-2ex] 
$(\osi{P}, \leq)$ & specified as input & the parent of link $i \in \os{L}_0$ \\
$(\osi{C}, \leq)$ & $\left( c: i \in \osi{P}[c] \right) $ & the children of link $i \in \os{L}_0$ \\
$(\osi{S}, \leq)$ & $\left( i \right) \cup \bigcup_{c \in \osi{C}} \osi{S}[c]$ & the
subtree rooted at link $i \in \os{L}_{-1}$ \\
$(\osi{A}, \geq)$ & $\left( i \right) \cup \bigcup_{p \in \osi{P}} \osi{A}[p]$ & the path from the root to link $i \in \os{L}_0$ \\
\end{tabular}
\end{center}
\caption{Connectivity sets of a kinematic tree as indexed ordered sets.  The
notation $(S, r)$ represents the set's name $S$ and its ordering relation $r$.}
\label{tab:kt}
\end{table}
There are several ways to represent the connectivity of a kinematic tree.  The
most common implementation is the mapping $j = p(i)$ from link $i$ to its
parent $j$ \cite{Featherstone2008}.  In Amplify, these mappings are implemented
as indexed ordered sets where the indices are the links in the tree.
Table~\ref{tab:kt} shows other choices for representing connectivity.  For
example, the kinematic tree can be represented as the map from link $i$ to its
$n$ children with the ordered set $\osi{C} = (c_1, c_2, \ldots, c_n)$ with
$c_1$ through $c_n$ representing other links in the tree.  These other ordered
sets can be derived in terms of the parent array, which Amplify parses from a
URDF-like representation of the $N$ robots. 

In the end, we rely on each of the representations in Table~\ref{tab:kt} when
implementing the CRBA, RNEA, and TJA as constraints.  The use of ordered sets is
partially motivated by the need to traverse the tree in the correct order.  For
example, the elements of the mass matrix $M(q)$ are computed starting from the
leaves of the tree to the root with the set $\Lk{-1}$.  Ordered sets also help
address an AMPL language requirement where decision variables in recursive
definitions cannot be referenced until they have been defined
\cite{Fourer2011}.


\subsection{The Composite Rigid Body Algorithm (CRBA)}
\begin{algorithm}[t]
\caption{The Composite Rigid Body Algorithm}
\label{alg:crba}
\begin{algorithmic}
\State Define composite spatial inertias and forces
\ConOne {$i \in \Lk{-1}$, $j \in \osi{C}$} $\spat[i]{I} = \spat[i]{I}[c] 
  + \sum_{j} \spat[j]{X}[i][T] \; \spat[j]{I} \; \spat[j]{X}[i]$
\Con {$i \in \Lk{-1}$, $j \in \osi{C}$, $k \in \osi{C}[j] \cap \osi{A}$}
  \State $\spat[j]{f}[i] = 
    \begin{cases}
      \spat[i]{I} \spats_i & i = j \\
      \spat[j]{X}[k] \spat[k]{f}[i] & \text{otherwise}
    \end{cases}$
\EndCon
\State Define $M(q)$
\Con {$i \in \Lk{-1}$, $j \in \Lk{-1}$}
  \State $M_{ij} = 
    \begin{cases}
      \spat[j]{f}[i] \spats_j & i \in \osi{S}[j] \\
      M_{ji} & j \in \osi{S}[i] \\
      0 & \text{otherwise},
    \end{cases}$
\EndCon
\end{algorithmic}
\end{algorithm}
\begin{lstlisting}[language=ampl,caption={The Composite Rigid Body Module},label=lst:mpm,float=t]
#----------- composite rigid body algorithm
var CRB_IC {i1 in SPAT_L[-1], i2 in SPAT_M, 
  i3 in SPAT_M, i4 in GRID[0] : i1 > 0} = 
    spat_I_im[i1,i2,i3] + 
      sum {i5 in SPAT_TREE_C[i1], i6 in SPAT_M, 
        i7 in SPAT_M} spat_X_ip[i5,i6,i2,i4] * 
          CRB_IC[i5,i6,i7,i4] * 
          spat_X_ip[i5,i7,i3,i4];

var CRB_f {i1 in SPAT_L[-1], i2 in SPAT_TREE_K[i1], i3 in SPAT_M, i4 in GRID[0] : i1 > 0} = 
  if i1 = i2 then 
    sum {i5 in SPAT_M} CRB_IC[i1,i3,i5,i4] * spat_s_ii[i1,i5] 
  else 
    sum {i5 in SPAT_M} spat_X_ip[prev(i2),i5,i3,i4] * CRB_f[i1,prev(i2),i5,i4];

var M {i1 in SPAT_L[-1], i2 in SPAT_L[-1], i3 in GRID[0] : i1 > 0 && i2 > 0} = 
  if i1 in SPAT_TREE_S[i2] then 
    sum {i5 in SPAT_M} CRB_f[i1,i2,i5,i3] * spat_s_ii[i2,i5] 
  else if i2 in SPAT_TREE_S[i1] then M[i2,i1,i3] 
  else 0;
\end{lstlisting}
The CRBA is an efficient algorithm for computing the inertia tensor $M(q)$ of a
robot (see Equations~\ref{eqn:M} and \ref{eqn:I}).  The algorithm can be derived
from differentiating the kinetic energy of the system in spatial coordinates,
substituting the joint velocities $\dot{q}$, and then differentiating with
respect to $\dot{q}$.  The set of constraints for solving $M(q)$ is shown in
Algorithm~\ref{alg:crba}.

In Algorithm~\ref{alg:crba}, the index $k \in \osi{C}[j] \cap \osi{A}[i]$ yields
the child of the $j^\text{th}$ link on the unique path from $i$ to the base.  A
careful ordering of $\osi{A}$ eliminates the need for explicitly defining $k$
(see Table~\ref{tab:kt}).  In our implementation, the value of $k$ is the
previous element relative to index $j$ in the set $\osi{A}$.  AMPL provides a
convenient shorthand for this operation with the \texttt{prev} operator
\cite{Fourer2011}.

Unlike implementation in an imperative programming language, the pseudocode of
Algorithm~\ref{alg:crba} is declaring a series of relationships.  In a modeling
language, the resulting code does not compute anything.  Finding values that
satisfy these constraints is the role of the solver.  Another difference is the
recursive definition of the algorithm.  When the lines of code are expanded,
these declarations are equivalent to the loop unrolling technique found in the
software library Pinocchio \cite{Carpentier2019}, where the body of the loop
is expanded until the loop is unnecessary.

Listing~\ref{lst:mpm} shows the implementation of the algorithm in AMPL.  The
listing also demonstrates how modularity is achieved with a set of variables and
constraints.  In general, the rest of the code in Amplify only depends on the
existence of the mass matrix $M(q)$ (see Figure~\ref{fig:mods}).  Variables such
as \texttt{CRB\_IC} are meant to be treated as local to the module.  A different
implementation of the module only has to ensure that $M(q)$ exists.  For
example, $M(q)$ can be defined in terms of the output of running the CRBA on
symbolic inputs \cite{Fevre2020,Ruscelli2022}.  The variables \texttt{CRB\_IC}
and \texttt{CRB\_f} can be safely deleted from the implementation without
affecting the rest of model.

\subsection{The Recursive Newton-Euler Algorithm (RNEA)}
The RNEA is an inverse dynamics algorithm, which given joint positions,
velocities, and accelerations, can calculate the joint torques in a recursively
efficient manner.  If the joint accelerations are zero, then the RNEA can
compute the vector of forces $b$ in Equation~\eqref{eqn:M}.  Solving for the
generalized force $b$ requires two passes of the tree.  The first pass
calculates the spatial velocities $\spatv_i$ and accelerations $\spata_i$ for
link $i$ in link $i$'s frame from the root of the tree to the leaves.  The
second pass calculates the net forces at each link from the leaves to the base.
The second pass also maps the spatial forces $\spatf_i$ back to the generalized
forces $b$.  The two passes are traditionally implemented with loops.  We
express the algorithm as a series of constraints in Algorithm~\ref{alg:rnea}.
\begin{algorithm}[t]
\caption{The Recursive Newton-Euler Algorithm}
\label{alg:rnea}
\begin{algorithmic}
\State Define spatial motions and forces of each link
\ConOne $\spatv_0 = 0$, $\spata_0 = -\spata_g$
\Con {$i \in \Lk{1}$, $p \in \osi{P}$}
  \State $\spatv_i = \spat[i]{X}[p] \spatv_p + \spats_i \dot{q}_i$
  \State $\spata_i = \spat[i]{X}[p] \spata_p + \spatv_i \times \spats_i \dot{q}_i$
\EndCon
\ConOne {$i \in \Lk{1}$} $\spat{\bar{f}}[i] = \spatI_i \spata_i + \spatv_i \times^* \spatI_i \spatv_i$ 
\ConOne {$i \in \Lk{-1} \setminus (0)$} $\spatf_i = \spat{\bar{f}}[i] + \sum_{c  \in \osi{C}} \spat[i]{X}[c][T] \spatf_c$
\State Define $b(q)$
\ConOne {$i \in \Lk{1}$} $b_i = \spats_i^T \spatf_i$
\end{algorithmic}
\end{algorithm}

\subsection{The Task Jacobian Algorithm (TJA)}
\label{ssec:tja}
The TJA of \cite{Featherstone2010b} serves as the basis for computing
scleronomic constraints of the form $y(q) = 0$, $\dot{y}(q) = 0$, and
$\ddot{y}(q) = 0$ with just the Amplify input data describing the robot and its
constraints.  The algorithm is designed to skip over the zeros in the
computation of Equation~\ref{eqn:rv} that result from the branch-induced
sparsity pattern of the kinematic tree.

The core algorithm and our extensions are expressed in Algorithm~\ref{alg:tja}.
The sets in the algorithm are defined in Tables~\ref{tab:kt}--\ref{tab:cons} and
the operator $[\cdot]_{rj}$ is defined in Equation~\ref{eqn:X}.  Our extensions
to the TJA computes the position $y^i(q)$ and velocity $\dot{y}^i(q)$,
transmission matrix $B^i(q)$, bias acceleration $\phi^i(q)$, and the constraint
Jacobian $J^i(q)$ of Equations~\ref{eqn:M}, \ref{eqn:I}, and \ref{eqn:v}.  The
extensions also assumes that a row of the constraint Jacobian $J^i(q)$ is
defined as a sum of $n_{J^K}$ spatial Jacobians of Equation~\ref{eqn:rv} such
that $J^i(q) = \sum_{k = 1}^{n_{J^K}} \spatr^{r}_k \spat[k]{J}(q)$.  This
property holds for the other quantities as well, e.g., $y^i(q)$ is equal to a
sum of different positions.  This provides Amplify the ability to constrain
links relative to each other.

In order to meaningfully sum these quantities, the user should choose to
orientate constraint frames to be axis aligned with respect to the root frame
$\{0\}$.  Based on user input for each constraint, we add and subtract positions
and velocities with a common alignment of the $x$-$y$-$z$ axes.  Our algorithm
labels these axes-aligned frames as $\{\bar{k}\}$.  The location of these frames
coincide with the location of the user-defined frame $\{k\}$ (see
Figure~\ref{fig:kt}).

When the constraint frame is axis aligned with the world frame, we internally
compute the rotation matrix $\spat[0]{E}[k]$ to obtain $\{\bar{k}\}$.  This
design choice simplifies the user's view of a constraint.  Through a pair of
inputs, a user specifies $\spat[k]{X}[b]$ and a binary input that specifies
whether the final frame is expressed with axes aligned with a world frame
$\{\bar{k}\}$ or body frame $\{k\}$.  Care should be taken that a summed
constraint is meaningful for the problem being solved.
\begin{table}[t]
\begin{center}
\scriptsize
\begin{tabular}{c|c|c}
\hline
Set & Definition & Description \\
\hline
$\us{R}$ & $\left\{ 1, \ldots, n_r \right\}$ & constraint IDs \\
\hline \\[-2ex]
$\usi{R}_r$ & $\{(k_1, b_1), (k_2, b_2), \ldots \}$ & constraint frames by row \\
\hline \\[-2ex]
$\csi{P}$ & specified as input & physical constraint IDs \\
$\csi{V}$ & specified as input & virtual constraint IDs \\
$\csi{U}$ & specified as input & control input IDs \\
$\csi{M}$ & $\csi{P} \cup \csi{V}$ & motion constraints \\
$\csi{F}$ & $\csi{P} \cup \csi{U}$ & force freedoms \\
$\csi{J}$ & $\csi{M} \cup \csi{F}$ & Jacobians \\
\hline \\[-2ex]
$(\Acon, \leq)$ & $\bigcup_{(k, b) \in \usi{R}_r} \osi{A}[b]$ & paths by phase \\
\end{tabular}
\end{center}
\caption{Sets of active constraints during phase $i \in \mathcal{P}$ and row $r
\in \csi{J}$ of $J^i$.  $\us{R}$ is independent of phase and $\Acon$ is
an ordered set.}
\label{tab:cons}
\end{table}
\begin{algorithm}[t]
\caption{The Task Jacobian Algorithm}
\label{alg:tja}
\begin{algorithmic}
\Con {$i \in \us{P}, r \in \csi{J}, (k,b) \in \usi{R}_r$}
  \State Define rotation from $\{j\}$ to root frame $\{0\}$
  \Con {$j \in \Acon,  \; p \in \osi{P}[j]$}
    \State $\spat[0]{E}[j] = \begin{cases}
    I & j = 0 \\
    \spat[0]{E}[p] \spat[j]{E}[p][T] & j > 0, \;
    \end{cases}$
    \State $\spatdot{0}{E}{j} = \begin{cases}
    0 & j = 0 \\
    \spat[0]{E}[j] {\spatv_j \times} & j > 0
    \end{cases}$
  \EndCon
  \State Define transforms from $\{b\}$ to $\{\bar{k}\}$ or $\{k\}$ 
  \If {$\{b\}$ to $\{k\}$}
    \State $\spat[\hat{k}]{X}[b] =  \spat[k]{X}[b]$
    \State $\spatdot{\hat{k}}{X}{b} = 0$
  \Else { $\{b\}$ to $\{\bar{k}\}$}
    \State $\spat[\hat{k}]{X}[b] =  \spat[0]{E}[b] \spat[b]{E}[k] \spat[k]{X}[b]$
    \State $\spatdot{\hat{k}}{X}{b} = \spatdot{0}{E}{b} \spat[b]{E}[k] \spat[k]{X}[b]$
  \EndIf
  \State Define transforms from $\{j\}$ to $\{b\}$
  %
  %
  \Con {$j \in \osi{A}[b]$}
    \State $\spat[b]{X}[j] = \begin{cases}
    I & b = j \\
    \spat[b]{X}[c] \spat[c]{X}[j] & b > j, \; c \in \osi{A}[b] \cap \osi{C}[j]
    \end{cases}$
    \State $\spatdot{b}{X}{j} = \begin{cases}
    0 & b = j \\
    {\left( \spat[b]{X}[j] \spatv_j - \spatv_b \right) \times} \spat[b]{X}[j] & b > j
    \end{cases}$
  \EndCon
  \State Define $J^i(q)$, $B^i(q)$, and $\phi^i(q, \dot{q}) = \dot{J}^i(q)\dot{q}$
  \Con {$j \in \osi{A}[b]$}
    \If {$r \in \csi{M}$}
      \State $[J^i]_{rj} = [J^i]_{rj}
         + \spatr_{\hat{k}} \; \spat[\hat{k}]{X}[{b}] \; \spat[b]{X}[j] \; \spat{s}[j]$
      \State $[\phi^i]_{r} = [\phi^i]_{r} + \spatr_{\hat{k}} \left(\spat[\hat{k}]{X}[{b}] \spatdot{b}{X}{j}  
         +  \spatdot{\hat{k}}{X}{b} \spat[b]{X}[j] \right) \spat{s}[j] \dot{q}_j$
    \ElsIf {$r \in \csi{F}$}
      \State $[B^i]_{rj} = [B^i]_{rj}
         + \spatr_{\hat{k}} \; \spat[\hat{k}]{X}[{b}] \; \spat[b]{X}[j] \; \spat{s}[j]$
    \EndIf
  \EndCon
  \State Define holonomic constraint $y^i(q)$
  \Con {$r \in \csi{M}$}
    \State $\spat[k]{X}[0] = \spat[k]{X}[b] \spat[b]{X}[0]$
    \State $[y^i]_r = [y^i]_r + \spatr_{\hat{k}} \begin{bmatrix}
           \ori(\spat[k]{X}[0]) \\ \pos(\spat[k]{X}[0]) 
           \end{bmatrix}$
  \EndCon
\EndCon
\end{algorithmic}
\end{algorithm}

\section{Specifying Constraints for a Class of Fixed-Step Runge-Kutta Methods}
\label{sec:ode}
We use Runge-Kutta methods to approximate solution trajectories of the
continuous ordinary differential equations (Equation~\ref{eqn:M}).  Runge-Kutta
methods are a family of one-step methods that iteratively compute the next point
of a continuous trajectory, say $q_{i+1}$ at time $t_{i+1}$, from the current
point $q_i$ at $t_i$ across $K$ stages.  Each stage computes the slope of the
trajectory evaluated at an interpolated point $q_j$ at $t_j$ that lies between
the current and next points with respect to time ($t_i \leq t_j \leq t_{i+1}$).
The next point $q_{i+1}$ is then a sum of the current point and a weighted
average of the $K$ derivative values.  For a one-degree-of-freedom system, we
can succintly write these equations as
\begin{equation}
\begin{aligned}
t_{i+1} &= t_i + h_i & t_j &= t_i + h_i \; c^K_j \\
q_{i+1} &= q_i + h_i \sum_{j = 1}^{K} b^K_j \dot{q}_j & q_j &= q_i + h_i \sum_{k
= 1}^{K} A^K_{jk} \dot{q}_k, \\
\end{aligned}
\label{eq:RK}
\end{equation}
where $A^K_{jk}$, $b^K_j$, and $c^K_j$ ($1 \leq j \leq K$) are the weights and
$h_i$ is the step size between $q_i$ and $q_{i+1}$.  With the appropriate
substitutions these equations also compute $\dot{q}_{i+1}$ from $\dot{q}_i$ and
$\ddot{q}_j$.  The extensions to degrees of freedom greater than one is also
straightforward.

Different Runge-Kutta methods arise based on the weights chosen to represent the
resulting approximation.  These weights are chosen to satisfy certain criteria,
such as satisfying a desired level of accuracy or computational performance
(e.g., first-same-as-last [FSAL] schemes).  A Butcher tableau \cite{Lubich2006,
Betts2010} is a common format for representing the weights of a Runge-Kutta
method:
\begin{equation}
B^K = \begin{array}{c|c}
c^K & A^K \\
\hline
    & (b^K)^T
\end{array} 
=
\begin{array}{c|ccc}
c^K_1  &  A^K_{11} & \ldots & A^K_{1K} \\
\vdots &  \vdots   & \ddots & \vdots   \\
c^K_K  &  A^K_{K1} & \ldots & A^K_{KK} \\
\hline
       &  b_1^K    & \ldots & b_K^K
\end{array},
\end{equation}
where $A^K_{jk}$, $b^K_j$, and $c^K_j$ of Equation~\ref{eq:RK} are elements of
the matrix $A^K$ and vectors $b^K$ and $c^K$, respectively.

In the end, implementing Equation~\ref{eq:RK} as constraints in a model and
requiring a Butcher tableau as input data removes the need to specify a
particular Runge-Kutta integration scheme.  An in-depth review of the different
types of Runge-Kutta methods are outside the scope of this article.  However,
certain classes of Runge-Kutta methods lead to duplicate constraints, like the
subset of FSAL schemes, where the constraints of the first and last stage used
to compute the state at grid point $i+1$ are equal to the constraints at grid
points $i$ and $i+1$, respectively.  Keeping duplicate constraints in an NLP can
lead to numerical issues or unnecessary variables and constraints.  We've
experienced each during earlier implementations of Amplify.

\subsection{Reducing the Variable Count}
In our current implementation, the number of grid points can be further reduced
by at most 2 if the first or last stage of a step results in the same
constraints as those at grid points $i$ and $i+1$, respectively.  The conditions
for eliminating these stages are
\begin{equation}
a_0 = \sum \limits_{j = 1}^K A^K_{1j}= 0, \quad a_K = \sum \limits_{j = 1}^K
A^K_{Kj} - b^K_j = 0,
\end{equation}
where the first stage is eliminated if $a_0 = 0$ and the last stage is dropped
if $a_K = 0$.  These conditions correspond to $q_1 = q_i$ and $q_K = q_{i+1}$ in
Equation~\ref{eq:RK}.  Both conditions are true for all FSAL integration schemes
such as Dormand-Prince and the family of Lobatto IIIA collocations methods
(including the trapezoidal and Hermite-Simpson methods).

\subsection{Expressivity and Flexibility in the NLP Integration Scheme}
In our implementation, the default spacing for time is uniform.  We assume that
the integration starts at $t_0$ and ends at $t_n$ with $h_i =
\frac{t_n-t_0}{n}$ for $0 \leq i \leq n$.  A developer or user can drop our
constraints on $h_i$ and implement their own step scheme.  In the end, the
variable $h_i$ is a free decision variable in Amplify that can be constrained
as needed. 

Furthermore, the integration scheme in Amplify can change at each grid point.
This type of flexibility provides opportunities for hybrid systems with
different phases (e.g., a hopper in the air vs.\ on the ground) or cascade-based
approaches \cite{Li2025a}.  In both these cases, a coarse integration scheme
can be chosen for phases of the trajectory that do not need the accuracy and a
finer grain integrator for areas that are expected to require it.  This type of
flexibility can be built into any library.  With our approach, the user defines
multiple Butcher tableaus and specifies the grid points they are associated
with.

\section{Specifying Constraints for Desired Trajectories using B\'{e}zier curves}
\label{sec:bez}
We use arbitrary-ordered B\'{e}zier curves for specifying desired trajectories
for inverse dynamics problems (e.g., applications of the hybrid zero dynamics
framework \cite{Westervelt2007}).  There are several methods for computing
B\'{e}zier curves, including the use of Bernstein polynomials, de Casteljau's
method, and matrix methods.  We use a matrix representation.  For those not
familiar with the matrix form of a B\'{e}zier curve, we start with a formulation
using Berstein polynomials and motivate the transition to a matrix formulation.
We conclude this section with a generic formula for computing an
$n^\text{th}$-order curve and its derivatives.

The typical representation of an $n^\text{th}$-order curve $\fun{b}{\R}{\R}$ in
the robotics trajectory optimization literature uses Bernstein polynomials
\cite{Fevre2019}:
\begin{equation}
b(\tau) = \sum_{i = 0}^n b_{i,n} P_i, \quad
b_{i,n} = \binom{n}{i} \tau^i (1 - \tau)^{n-i}
\label{eq:bern}
\end{equation}
where $\tau \in [0, 1]$ is the input, $\binom{n}{i}$ is a binomial coefficient,
$P_i \in \R$ is a B\'{e}zier control point, and $b_{i,n}$ is a Berstein basis
polynomial of degree $n$.  While straightforward to code, a downside of
Equation~\ref{eq:bern} is the lack of computational reuse.  The only values that
change when solving an OP that makes use of Equation~\ref{eq:bern} are $\tau_i$
and $P_i$.  A matrix representation allows for greater computational reuse by
collecting most of the constants into a B\'{e}zier matrix $B$.  Letting
$\pmb{\tau} = [1, \tau, \ldots, \tau^n ]^T$, $P = [P_0, \ldots, P_n]$, and
applying the relation $(1 - t)^{n-i} = \sum_{k = 0}^{n - i} \binom{n - i}{k}
(-1)^k t^k$, Equation~\ref{eq:bern} can be rewritten as
$b(\tau) = \pmb{\tau}^T B P$, where
\begin{equation}
[B]_{rc} = \begin{cases}
\binom{n}{c} \binom{n-c}{r-c} (-1)^{r-c} & c \leq r \\
0 & \text{otherwise},
\end{cases}
\label{eq:bezmat}
\end{equation}
$0 \leq r \leq n$ and $0 \leq c \leq n$.\footnote{The pair of indices $(r, c)$
and $(i, k)$ are related: $i = c$ and $k = r - c$}  The product of the binomial
coefficients in Equation~\ref{eq:bezmat} can be further simplified:
\begin{equation}
\begin{gathered}
\binom{n}{c} \binom{n-c}{r-c} = \frac{n^{\underline{r}}}{c!(r-c)!} = \prod_{j = 0}^{r-1} \frac{n-j}{a(j, c) a(j, r - c)} \\
a(j, k) = k - \min(j, k - 1),
\end{gathered}
\end{equation}
where $n^{\underline{r}} = \prod_{j = 0}^{r-1} (n - j)$ is the falling factorial.

Taking advantage of the fact that the derivatives of a B\'{e}zier curve is
another B\'{e}zier curve, we can compute an $n^\text{th}$-order B\'{e}zier curve
and its $d$ derivatives as
\begin{equation}
\sd{^db}{\tau^d} = \sum_{r = d}^n \sum_{c = 0}^r r^{\underline{d}} \; \tau^{r-d}
B_{rc} P_c.
\label{eqn:bezd}
\end{equation}
The indexing skips over known zero elements in $\sd{^d\pmb{\tau}}{\tau^d}$ and $B$.

Our motivation for introducing a polynomial trajectory module is to serve as a
replacement library for hybrid zero dynamics research.  We compare our
implementation with \cite{Fevre2020} in the examples section.

\section{Examples}
\label{sec:ex}
We have tested Amplify on 7 different robot systems commonly used as benchmark
problems: the acrobot, a cart-pole system, a moving block, a kinematic car, an
object with contact points for grasping, RABBIT, and Spot.  Five of the problems
are demos reimplemented in Amplify from three NLP libraries: OptimTraj
\cite{Kelly2017}, TROPIC \cite{Fevre2020}, and Horizon
\cite{Ruscelli2022}.  These libraries were chosen because it was possible to
understand the innerworkings of the code, especially when it came to extracting
the objective, variables, and constraints of the problem being duplicated.  We
provide further background on each library in the Appendix.

A summary of performance results are shown in Tables~\ref{tab:oj}--\ref{tab:hn}.
The data in these tables compare Amplify and the other NLP libraries in terms of
the optimal value of the objective, number of iterations, elapsed wall time, and
problem size averaged over 10 runs for each demo rounded to the nearest third
decimal place.  Objective values within a demo are always directly comparable
between Amplify and the other library.  However, keep in mind that the numbers
are not always an apples-to-apples comparison as details vary across solvers and
release versions.  We highlight known instances when analyzing the results in
the following subsections.

Finally, in terms of timing, we measure elapsed time from when a demo script
starts, including problem setup, to when a call to the solver returns back to
the script.  To more accurately capture computation time, we lightly modified
the OptimTraj, TROPIC, and Horizon demo scripts to minimize printing
intermediate results to screen.  All data and tools used to create the tables
are available online.

\subsection{Comparison of Select Demos from OptimTraj's Library}
\begin{table*}[t]
\centering
\scriptsize
\caption{Comparison Between OptimTraj + fmincon v24.1 and Amplify + Ipopt
3.12.13 across 10 runs. Better performance is highlighted in gray. TRAP and HS
correspond to the trapezoidal and Hermite-Simpson direct collocation methods,
respectively.}
\label{tab:oj}

\begin{tabular}{llccccccc}
\toprule
Demo & Library & RK Method & Objective & Wall Time (s) & Iterates & \# Vars. & \# Eq. Cons. & \# of Ineq. Constr. \\
\midrule

\multirow{2}{*}{Acrobot}
 & OptimTraj & TRAP & \cellcolor{gray!20}\textbf{38.96} & 0.65 $\pm$ 0.50 & 55 & 102 & 76 & 102 \\
 & Amplify   & TRAP & 69.88 & \cellcolor{gray!20}\textbf{0.23} $\pm$ 0.03 & \cellcolor{gray!20}\textbf{14} & 132 & 116 & 18 \\

\midrule

\multirow{2}{*}{Cart-pole (force)}
 & OptimTraj & TRAP & 229.29 & 0.43 $\pm$ 0.02 & 38 & 152 & 116 & 152 \\
 & Amplify   & TRAP & 229.29 & \cellcolor{gray!20}\textbf{0.23} $\pm$ 0.02 & \cellcolor{gray!20}\textbf{12} & 202 & 176 & 0 \\

\midrule

\multirow{2}{*}{Cart-Pole (time)}
 & OptimTraj & TRAP & 1.15 & 0.46 $\pm$ 0.05 & 158 & 52 & 36 & 52 \\
 & Amplify   & TRAP & 1.15 & \cellcolor{gray!20}\textbf{0.22} $\pm$ 0.02 & \cellcolor{gray!20}\textbf{30} & 80 & 74 & 0 \\

\midrule

\multirow{2}{*}{Moving Block}
 & OptimTraj & TRAP & 12.06 & \cellcolor{gray!20}\textbf{0.06} $\pm$ 0.02 & 19 & 92 & 58 & 92 \\
 & Amplify   & TRAP & 12.06 & 0.18 $\pm$ 0.02 & \cellcolor{gray!20}\textbf{7} & 116 & 88 & 0 \\

\midrule

\multirow{2}{*}{Five-Link Biped (5 DOF)}
 & OptimTraj & HS & 420.13 & 0.59 $\pm$ 0.10 & 27 & 197 & 132 & 225 \\
 & Amplify   & HS & \cellcolor{gray!20}\textbf{260.43} & \cellcolor{gray!20}\textbf{0.50} $\pm$ 0.08 & \cellcolor{gray!20}\textbf{11} & 266 & 212 & 2 \\

\midrule

Five-Link Biped (7 DOF)
 & Amplify & HS & 443.14 & 0.53 $\pm$ 0.03 & 13 & 326 & 300 & 2 \\

\bottomrule
\end{tabular}
\end{table*}
We ported the acrobot, cart-pole, moving-block, and RABBIT demos from the
OptimTraj library.  A summary of the tasks and objective functions of each demo
is given below:
\begin{description}
\item[Acrobot:] Balance a double-pendulum from a downward equilibrium position
to an upright equilibrium position in a fixed amount of time.  The objective is
to minimize the sum of the squared torques.

\item[Cart-Pole:] Balance a pendulum affixed to a cart at its upward equilibrium
position starting from the downward equilibrium position.  The cart moves on a
fixed track.  This problem has two separate objectives, where it either achieves
swing up in 1) a fixed amount of time while minimizing the sum of the squared
forces or 2) the shortest amount of time possible.

\item[Moving Block:] Move a block from one position at rest to another in a
fixed amount of time.  The objective is to minimize the sum of the squared input
force.

\item[RABBIT:] Find a periodic motion of a symmetric biped robot using an
open-loop control scheme.  The objective is to minimize the sum of the input
torques squared.

\end{description}

The values for each of the physical parameters and initial seed values are taken
from their respective demos.  Amplify uses Ipopt's and OptimTraj uses fmincon's
interior-point solvers at version 3.12.13 and 24.1 (R2024a), respectively.  We
left each solver at their default options as set by Ipopt and OptimTraj,
respectively.

The results are summarized in Table~\ref{tab:oj}.  When comparing results, the
optimal value of the objective can be directly compared within implementations
of each demo.  The optimal values are effectively the same for the cart-pole ($J
\approx \qty{229.29}{N^2}$ for minimal force and \qty{1.15}{s} for minimal time)
and moving-block ($J \approx \qty{12.06}{N^2}$) demos.  OptimTraj + fmincon
performs better on the acrobot problem.  Each of these demos use the trapezoidal
(TRAP) RK method for integrating the equations of motion.

With RABBIT, the demo implementation in OptimTraj solves for the optimal
open-loop actuation over a step using the model of the five-link biped RABBIT
over 6 grid points.  There are five integration schemes available to choose
from.  We chose their recommended integration scheme for the problem, which was
the Hermite-Simpson (HS) direct collocation method with gradients of the
objective and constraint functions computed analytically.

In our port, we implemented two different versions of the demo in Amplify.  For
each version, the coordinate systems used in OptimTraj and Amplify are
different.  In our 5-DOF model, the coordinate systems are a different set of
minimal coordinates.  In order to compare the cost of each trajectory, we
applied a coordinate transformation from OptimTraj's solution into our minimal
coordinate system.  After mapping the cost of OptimTraj's trajectory into our
coordinate system, we see that Amplify + Ipopt performs better ($J \approx
\qty{260.43}{N^2 m^2}$ vs.\ $J \approx \qty{420.13}{N^2 m^2}$, respectively).
Our OptimTraj results are consistent with the published results in
\cite{Kelly2017}.

We also implemented the RABBIT model using a 7-DOF floating-base model in
Amplify.  The motivation was to test an assertion in \cite{Kelly2017} that
constrained systems are harder to solve as NLPs.  We did not encounter any
issues with this model.  The cost is the highest of the three implementations
($J \approx \qty{443.14}{N^2 m^2}$).  This is likely due to the addition of a
pair of physical holonomic constraints that pin the foot of the stance leg to
the ground creating a pivot joint.  Relative to our implementation in minimal
coordinates, adding a pair of PHCs increased the problem size by 60 decision
variables and 88 constraints.  It has comparable performance metrics to the
Amplify 5-DOF model despite the larger problem size.

When comparing timing performance, the wall time gives an idea of which process,
on average, took longer to generate a model and converge to an optimal
trajectory.  The wall time also includes time the OS is busy performing other
tasks.  This is a major reason we modified all demos to not print information to
screen while the NLP runs.  AMPL does provide more granular timers that keep
track of the amount of time AMPL and a solver spend running user-level and
system-level tasks on a CPU.  It also reports memory usage, which is a useful
number for resource-constrained devices.  Our online material logs the
additional timing data.  We do not log memory usage.

Table~\ref{tab:oj} also lists the numbers of iterations and problem size.  These
numbers should be used for informational purposes only.  For example, each
solver has a different notion of what constitutes an iteration.  In general, we
expect more iterations and longer run times associated with OptimTraj due to
Matlab's computational speed as a general-purpose language and fmincon's
documented slower execution speed when compared to Ipopt (e.g., \cite{Lee2016,
Yamamoto2025}).

\subsection{Comparison of 7-DOF Biped Demo from TROPIC's Library}
\begin{table*}[t]
\centering
\scriptsize
\caption{Comparison between TROPIC + Ipopt 3.12.3 + spat\_v2 and Amplify + Ipopt
3.12.13 across 10 runs of TROPIC's five-link biped NLP. The RK method is Hermite
Simpson.  Better performance is highlighted in gray.}
\label{tab:tc}

\begin{tabular}{lcccccccc}
\toprule
Library & Objective & Wall Time (s) & Iterates & Eval. Time (s) & \# Vars. & \# Eq. Cons. & \# Ineq. Cons. & Note \\
\midrule

TROPIC 
& \cellcolor{gray!20}\textbf{4.12} 
& \cellcolor{gray!20}\textbf{2.69 $\pm$ 0.41} 
& \cellcolor{gray!20}\textbf{101} 
& 1.15 $\pm$ 0.17 
& 1404 & 6 & 1642 
& TROPIC defaults with $\epsilon$ constraints \\

TROPIC 
& 6.09 
& 18.97 $\pm$ 0.44 
& 793 
& 9.20 $\pm$ 0.21 
& 1404 & 1390 & 258 
& TROPIC defaults with equality constraints \\

TROPIC 
& 5.92 
& 207.48 $\pm$ 2.46 
& 1141 
& 69.76 $\pm$ 0.80 
& 1404 & 1390 & 258 
& Ipopt defaults with equality constraints \\

Amplify 
& 5.73 
& 31.95 $\pm$ 0.63 
& 206 
& 11.90 $\pm$ 0.29 
& 2345 & 2331 & 206 
& Ipopt defaults with equality constraints \\

\bottomrule
\end{tabular}
\end{table*}
\begin{table*}[t]
\centering
\scriptsize
\caption{Comparison Between Horizon + Ipopt 3.12.3 and Amplify + Ipopt 3.12.13
on an 18-DOF Spot quadruped model across 10 runs. Gray highlights better
performance, dashes denote solver failed to converge.}
\label{tab:hn}

\begin{tabular}{lcccccccc}
\toprule
Library & RK Method & Objective & Wall Time (s) & Iterates & Eval. Time (s) & \# Vars. & \# Eq. Cons. & \# Ineq. Cons. \\
\midrule

Horizon  & \multirow{2}{*}{COL3} & - & - & - & - & - & - & - \\
Amplify  & 
& \cellcolor{gray!20}\textbf{13250.85} 
& \cellcolor{gray!20}\textbf{199.74 $\pm$ 2.11} 
& \cellcolor{gray!20}\textbf{172} 
& 152.02 $\pm$ 1.20 
& 14854 & 14228 & 480 \\

\midrule

Horizon  & \multirow{2}{*}{RK1} & - & - & - & - & - & - & - \\
Amplify  & 
& \cellcolor{gray!20}\textbf{13087.32} 
& \cellcolor{gray!20}\textbf{217.56 $\pm$ 4.39} 
& \cellcolor{gray!20}\textbf{201} 
& 168.97 $\pm$ 3.84 
& 3706 & 3080 & 480 \\

\midrule

Horizon  & \multirow{2}{*}{RK4} 
& \cellcolor{gray!20}\textbf{13127.37} 
& \cellcolor{gray!20}\textbf{43.20 $\pm$ 0.85} 
& 330 
& 9.12 $\pm$ 0.26 
& 3400 & 2787 & 1180 \\
Amplify  & 
& 13343.08 
& 159.83 $\pm$ 2.95 
& \cellcolor{gray!20}\textbf{136} 
& 119.61 $\pm$ 2.46 
& 14836 & 14210 & 480 \\

\bottomrule
\end{tabular}
\end{table*}
With TROPIC, we are able to perform a more direct comparison as both libraries
use Ipopt as the underlying solver and rigid body dynamics algorithms to compute
the equations of motion.  We compare against the 7-degree-of-freedom (DOF) biped
demo script found in the library's online code repository.  The demo code solves
a closed-loop, optimally-actuated step using the hybrid zero dynamics framework
on a custom five-link model of a planar biped over 25 grid points.  The
equations of motion have both physical and virtual holonomic constraints that
need to be satisfied at each grid point.  The objective is to minimize the sum
of the squared joint torques.  The use of the same solver at the v3.12 release
branch and biped model enables a direct comparison of the transcription
approaches used in each library.

In terms of results, when the TROPIC library and demo script are left at their
default settings, TROPIC is able to find an optimal trajectory with a cost of
$\qty{4.12}{N^2 m^2}$ in $\qty{2.69}{s}$.  The convergence time is close to
published results of \qty{1.50}{s} \cite{Fevre2020}.  Amplify is able to find an
optimal trajectory with a cost of $\qty{5.73}{N^2 m^2}$ in $\qty{31.95}{s}$.
The times are the reported Ipopt solution times started from the same seed
value.  However, upon further inspection, we noticed that there are important
details in the implementation of TROPIC that we do not make in Amplify.

TROPIC represents equality constraints $c(z) = 0$ as inequality constraints of
the form $-\epsilon \leq c(z) \leq \epsilon$ with $\epsilon > 0$ and of
appropriate dimension.  There are strong theoretical and practical reasons,
especially for interior-point methods, why this is not a recommended design
pattern (see Ipopt documentation).  For interior-point methods, this results in
the equality constraint not being satisfied to the internal tolerances that a
solver like Ipopt would have achieved otherwise, unnecessarily increases the
problem size, and can lead to numerical and convergence issues.

In particular, we noticed that playing around with TROPIC's $\epsilon$
controlled the convergence speed to an optimal trajectory from the given seed
value.  The parameter is explicitly set in the demo script making it a parameter
that the user tunes on their own.

Table~\ref{tab:tc} benchmarks the two libraries.  The multiple TROPIC entries
correspond to running the NLP under different options set in TROPIC.  These
modifications are detailed in the notes column.  For example, an equivalent
formulation of TROPIC that mimics how we represent equality constraints in AMPL
is to set $\epsilon = 0$ (TROPIC defaults with equality constraints).  In this
case, Ipopt converges to a trajectory with a cost of $\qty{6.09}{N^2 m^2}$ in
$\qty{18.97}{s}$.  Furthermore, if equality constraints are still set with
$\epsilon = 0$ and the Ipopt options are left at their default values (Ipopt
defaults with equality constraints), the cost improves to $\qty{5.92}{N^2 m^2}$
but TROPIC takes $\qty{207.48}{s}$ to find an optimal solution.  In summary, the
performance achieved in TROPIC are the results of tuned TROPIC-specific and
Ipopt-specific options (e.g., \texttt{IPOPToptions.m} in TROPIC's repository).
Modifying these options leads to major changes in performance for this demo.

\subsection{Comparison of 18-DOF Quadruped Demo from Horizon's Library}
Horizon is another library that provides a direct comparison with Amplify.  The
library uses CasADi to compute derivatives of objective and constraint
functions.  It uses Pinocchio to compute quantities like generalized inputs $u$,
constraint forces $f$, and accelerations $\ddot{q}$ of Equation~\ref{eqn:M} with
spatial vectors and rigid body dynamics algorithms.  In particular, the
quadruped demo uses a variant of the RNEA to compute input forces.

In the results published in \cite{Ruscelli2022}, the quadruped Spot has to jump
from an initial position at rest to a final position at rest but rotated by
$120^\circ$.  The robot model has 12 internal DOFs attached to a floating base.
The control inputs are accelerations $\ddot{q}(t)$ and ground contact forces at
each foot $f(t) \in \R^{12}$.  The cost function is $J(\cdot) = 3 \sum_{i = 0}^N
\dot{q}_i^T\dot{q}_i + 0.02 \sum_{i = 0}^{N-1} f_i^Tf_i$ ($0 \leq i \leq N$),
where $\dot{q}_i$ and $f_i$ are the values of $\dot{q}$ and $f$ evaluated at a
grid point $i$.

We modified the demo script to generate an NLP with respect to three of the RK
methods Horizon implements.  These are the implicit $5^\text{th}$-order
Gauss-Legendre collocation (COL3), explicit Euler (RK1), and classical explicit
$4^\text{th}$-order (RK4) methods.

In terms of performance (Table~\ref{tab:hn}), Amplify is able to achieve a
similar cost for the jumping motion while satisfying the same objective and
constraints as the Horizon demo code.  Amplify converges to an optimal solution
in all 3 cases, but Horizon only converges for the RK4 instance.  However, when
Horizon converges, it does so much faster than Amplify.

When we take into consideration that Amplify converges in fewer iterations than
Horizon, we suspect the difference in speed is due to two important factors
(IF):

\begin{enumerate}[label=IF\arabic*]
\item \label{enum:hand} the human hand tuning in reducing the number of NLP
variables and constraints in Horizon outperforms AMPL's presolver, and

\item \label{enum:casa} CasADi generates more performant code than AMPL when it
comes to evaluating the objective and constraint functions for the Spot NLP.

\end{enumerate}
With respect to \ref{enum:hand}, the overall implementation of the Horizon
library and its demos makes use of several trajectory optimization modeling
tricks to reduce variable and constraint count.  For example, the RK4
implementation uses a compressed form of the constraints to eliminate variables
and constraints at the internal $K$ stages \cite{Betts2010}.  This alone
reduces the problem size by nearly a factor of $K = 4$ relative to Amplify.
However, the library has difficulty converging with other ODE solvers.  We
attempted minor changes to the demos that failed to converge, like relaxing the
demo-specific tolerance settings for Ipopt.  Ipopt did not converge with these
changes either.

For \ref{enum:casa}, CasADi and AMPL generate expression graphs of the
objectives and constraints.  They take different approaches in compiling the
graph and the related derivatives \cite{Andersson2018}.  Given that solvers
compute these quantities many times per iteration, fast runtime and low overhead
is key.  The order of magnitude faster evaluation time in the RK4 version of the
Spot NLP of Table~\ref{tab:hn} clearly shows that CasADi's expression graph is
better optimized for this particular case.

While CasADi has been shown to outperform AMPL in simple benchmark problems
\cite{Andersson2012}, it did not outperform AMPL as a backend in our
comparison with TROPIC as performance tuned features were turned off.  However,
the Horizon results suggest that our library may not scale well with DOFs.  This
might be due to the AMPL backend not generating performant enough code relative
to CasADi (see Eval. Times).  It could also be that our formulation limits the
assumptions the presolver can make when eliminating variables and constraints.
In the end, we leave it as future work to better isolate the difference in
speed.  While we are able to replicate the Spot demo, the NLPs are not exact
duplicates.  Furthermore, our library has not been as heavily optimized as
Horizon, Pinocchio, and CasADi.  A key limitation is the closed-source nature of
AMPL.

Finally, reproducing the results in \cite{Ruscelli2022} is not straightforward.
We document issues in \cite{Rosa2026a}. 

\subsection{Time-Optimal Motion of a Kinematic Car}
\begin{figure}[t]
\centering

\newlength{\colheight}
\setlength{\colheight}{0.35\textheight}

\begin{minipage}[c]{0.38\columnwidth}
    \centering
    \subfloat[An optimal motion]{%
        \includegraphics[
            width=\linewidth,
            height=\colheight,
            keepaspectratio
        ]{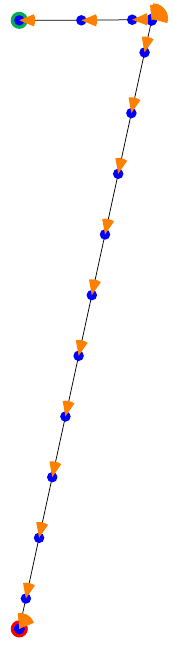}
    }
\end{minipage}
\hspace{-0.06\columnwidth}
\begin{minipage}[c]{0.63\columnwidth}
    \centering

    \subfloat[Ipopt and conopt]{%
        \includegraphics[
            width=\linewidth,
            height=0.49\colheight,
            keepaspectratio
        ]{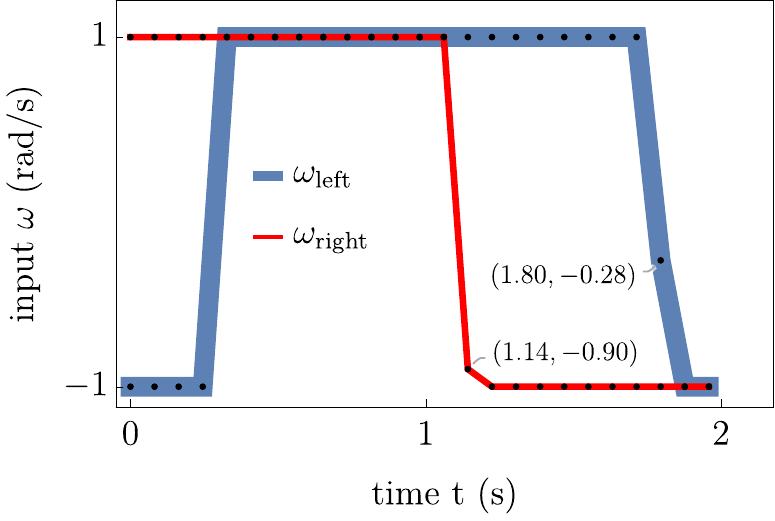}
    }

    \vfill

    \subfloat[Gurobi and LOQO]{%
        \includegraphics[
            width=\linewidth,
            height=0.49\colheight,
            keepaspectratio
        ]{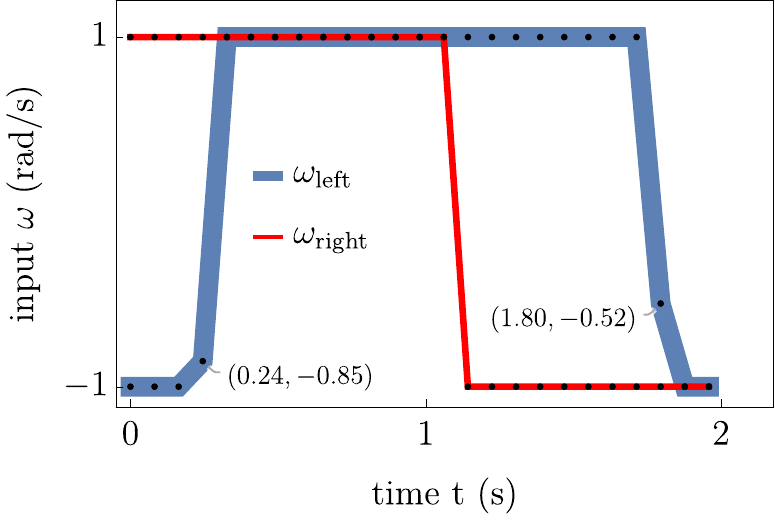}
    }

\end{minipage}

\caption{(a) A time-optimal motion for a differential-drive kinematic car (blue
dot) and its heading (orange cone) from its start (red circle) to its goal
(green circle) configuration.  The optimal motion is to turn left, drive
forward, turn right, and then drive backwards.  Only 4 of the 9 solvers tested
converged to a time-optimal solution of $t \approx \qty{1.96}{s}$. (b) and (c)
the control inputs applied by the solvers.  The black dots are the grid points
where the inputs are applied.}

\label{fig:car}
\end{figure}
We present results on finding time-optimal trajectories for a differential drive
robot with bounds on its inputs.  The problem has been fully solved through
algorithmic means in \cite{Balkcom2000}.  We take our model parameters from
\cite{LaValle2009}.  The task is to go from a start configuration to the origin
in minimal time.  The control inputs are the angular velocities of the left and
right wheel of the robot, $\omega_\text{left}$ and $\omega_\text{right}$,
respectively.  The inputs are bounded at $\pm \qty{1}{rad/s}$.

In general, time-optimal problems with bounds on the controls are challenging
benchmarks for trajectory optimization problems.  Most solutions require the use
of a bang-bang controller that switches between the values of zero and the upper
and lower bounds of the actuator limits at discrete instances in time.  These
discrete switches can make it difficult for gradient-based solvers to accurately
approximate the optimal solution as the derivatives become nonsmooth in the
control inputs.  Being tied to a single solver can be a limitation if the NLP
does not converge or the optimal trajectory is not satisfactory.

In this example, we present the results of solving an NLP version of the time
optimal problem with the nine AMPL-compatible nonlinear constrained optimization
solvers: conopt, filter, Gurobi, Ipopt, Knitro, Lancelot, LOQO, MINOS, and
Snopt.  In our tests, only conopt, Gurobi, Ipopt, and LOQO converged to a
solution.  Averaged over 10 trials, they each had wall times in seconds of 0.35
$\pm$ 0.03, 3.85 $\pm$ 0.09, 0.94 $\pm$ 0.06, and 0.30 $\pm$ 0.18, respectively.
These trials always yielded an objective value of 1.96 solved over 391 decision
variables and 341 equality constraints.  Figure~\ref{fig:car} shows the
trajectory.

\subsection{Four-Fingered Grasp of a Block}
\begin{figure*}[t]
  \centering

  \subfloat[Gurobi]{%
      \includegraphics[width=0.24\textwidth]{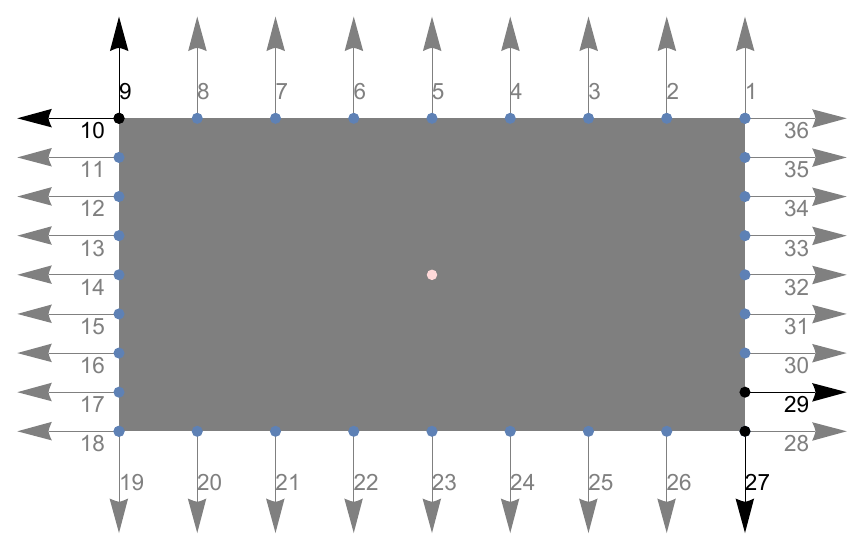}
      \label{fig:gg}
  }
  \subfloat[Knitro and RAPOSa]{%
      \includegraphics[width=0.24\textwidth]{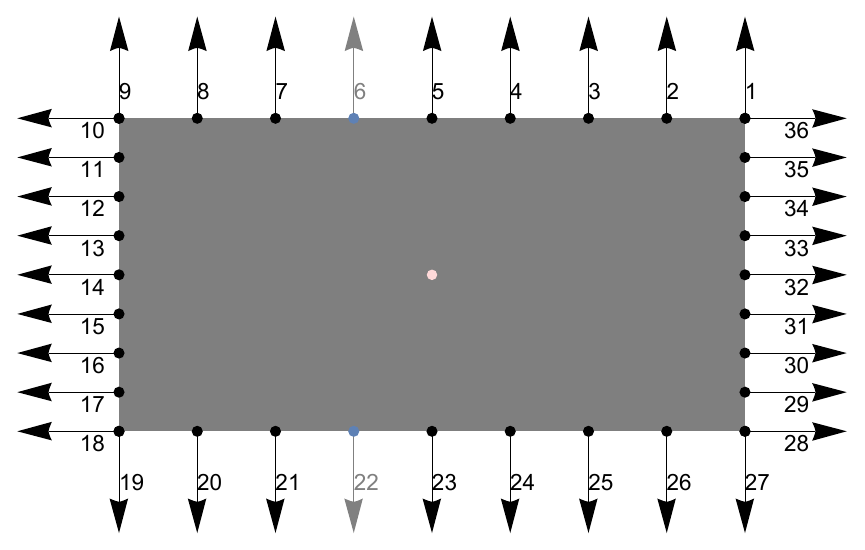}
      \label{fig:gkr}
  }
  \subfloat[conopt]{%
      \includegraphics[width=0.24\textwidth]{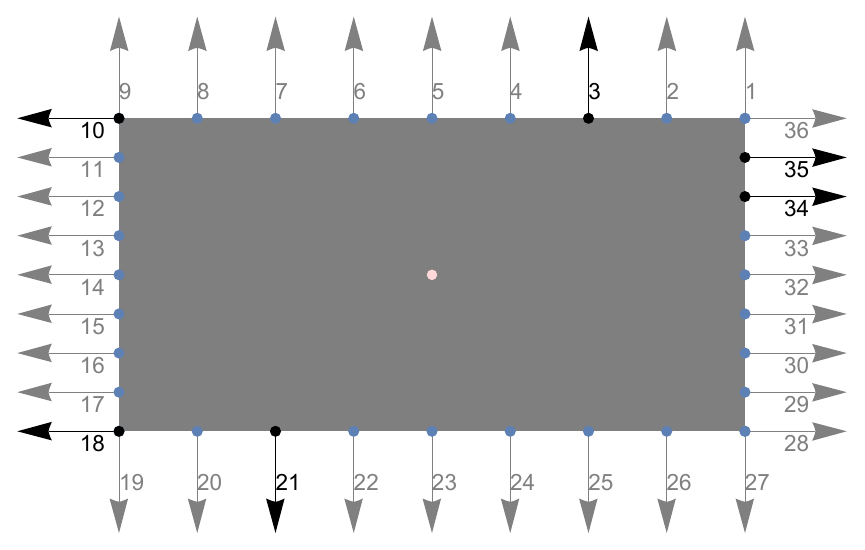}
      \label{fig:gc}
  }
  \subfloat[scip]{%
      \includegraphics[width=0.24\textwidth]{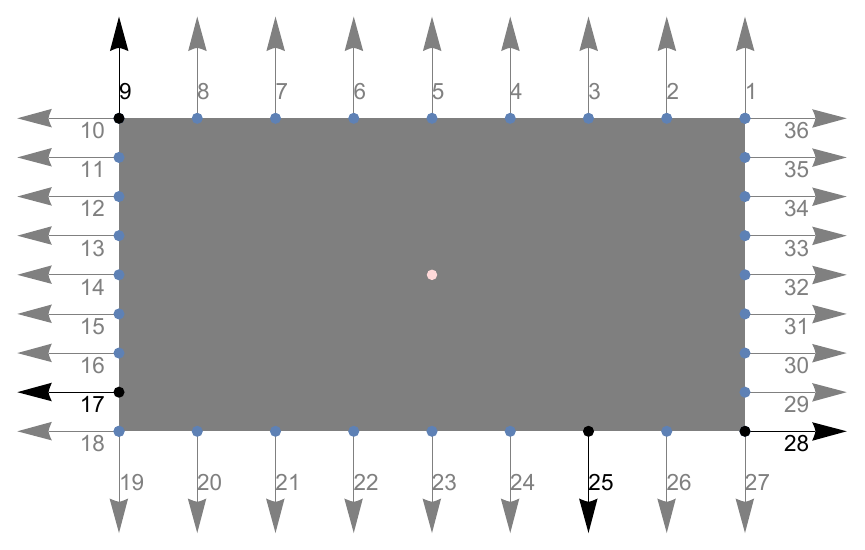}
      \label{fig:gs}
  }

  \caption{Select solver results of the best minimal fingered form-closure
grasp.  The optimal solution, e.g., (a) and (d), will always use 4 frictionless
point-contact fingers each placed on a side of the block as depicted by the
object's normals (black arrows).  The problem is formulated as a mixed-integer
program.  Grasp points are shown in black.  RAPOSa used Knitro as an underlying
solver, which is why it found the same solution as Knitro.}

  \label{fig:grasp}
\end{figure*}
In this final example, we demonstrate the ability of formulating objectives and
constraints with binary and integer decision variables.  The task is to minimize
the number of fingers needed to place a rectangular planar block in form
closure.  The fingers are modeled as frictionless point contacts (FPC).  They
can be placed at 36 discrete points on the block (see Figure~\ref{fig:grasp}).
The number of fingers that can contact the block is between 4 and 36.  The
optimal number is 4 \cite{Lynch2023}.  To satisfy the form closure constraint, a
finger has to be placed on each side of the block.  In total, there are
approximately 68.7 billion possibilities.  While this search space is large,
there are only $\binom{36}{4} = 58,905$ different ways to place 4 fingers on a
block of which $2,592$ are in form closure.

In our NLP model, we use physical holonomic constraints to discretize the
contact points on the block and a binary vector of decision variables to keep
track of which contact points are active.  When coupled with constraints $q =
\dot{q} = \ddot{q} = 0$, Equation~\ref{eqn:M} simplifies to $J^T_p(0)f = 0$ for
$f \geq 0$, i.e., the form-closure constraint \cite{Lynch2023}.  In this
context, $J^T_p(0)$ is the grasp map \cite{Murray1994}.  Listing~\ref{lst:mip}
is an example of how a user could implement the NLP model on top of Amplify.  
\begin{lstlisting}[language=ampl,caption={An example mixed integer
problem.},label=lst:mip,float=t]
set SURFACE := {'n', 't', 'u'};
param rot {SURFACE} >= 0;
param pos {SURFACE} >= 0;

# contact points and fingers
param nc; # num contacts
param nf; # num fingers

param mu; # linear friction
param gamma; # angular friction

#-- object only (no fingers modeled as rigid bodies)
set CONTACTS := {i1 in R : 
  i1 <= nc*nm and i1 mod nm = 1};

var c {CONTACTS} binary; # in contact?

#-- zero unused contact forces
subject to FC_NO_CONTACT {i1 in CONTACTS, 
  i2 in SPAT_M, i3 in GRID[0]: i2 = pos['n']} : 
    (1 - c[i1]) * f[1,i1+i2-1,i3] = 0;

subject to FC_NO_WRENCH {i1 in CONTACTS, 
  i2 in SPAT_M, i3 in GRID[0]: i2 != pos['n']} : 
    f[1,i1+i2-1,i3] = 0;

#-- FPC friction model
subject to FPC {i1 in CONTACTS, i2 in GRID[0]} : 
  f[1,i1+pos['n']-1,i2] >= 0;

#-- statics
subject to Q_OBJECT 
  {i1 in SPAT_M, i2 in GRID[0]} : q[i1,i2] = 0;
subject to V_OBJECT 
  {i1 in SPAT_M, i2 in GRID[0]} : v[i1,i2] = 0;
subject to A_OBJECT 
  {i1 in SPAT_M, i2 in GRID[0]} : a[i1,i2] = 0;

#-- form-closure constraints
param tol_det;
set WRENCH;

var G {i1 in WRENCH, i2 in WRENCH, i3 in GRID[0]} = 
  sum {i4 in CONTACTS} J[1,i4+pos['n']-1,i1,i3]
      *c[i4]*J[1,i4+pos['n']-1,i2,i3];

var det_G {i1 in GRID[0]} = 
  G[1,1,i1]*(G[3,3,i1]*G[5,5,i1]-G[3,5,i1]^2) - 
  G[1,3,i1]*(G[1,3,i1]*G[5,5,i1]
    -G[1,5,i1]*G[3,5,i1]) + 
  G[1,5,i1]*(G[1,3,i1]*G[3,5,i1]
    -G[1,5,i1]*G[3,3,i1]);

subject to MATRIX_RANK {i1 in GRID[0]} : 
  tol_det <= det_G[i1];

subject to FORM_CLOSURE {i1 in CONTACTS, 
  i2 in GRID[0]} : f[1,i1+pos['n']-1,i2] >= c[i1];

check {i1 in SPAT_M} : spat_ag[i1] = 0;
  # requires zero gravity as input

#-- solve a minimal fingers problem
var fingers >= nf, <= nc integer;
subject to NUM_CONTACTS : 
  sum {i1 in CONTACTS} c[i1] = fingers;
minimize NUM_FINGERS : fingers;
\end{lstlisting}

We ran this example against several mixed-integer solvers.  The results are
visualized in Figure~\ref{fig:grasp}.  To get these results, we had to change
the AMPL presolver's default behavior as it kept terminating our script with
claims that the problem was infeasible.  Otherwise, we left all solvers at their
default values.  While the goal of this example is to demonstrate a
mixed-integer program, we conjecture that some solver-related option tuning is
necessary.  However, option tuning is outside the scope of this paper.

\section{Discussion}
\label{sec:dis}
When referring back to Table~\ref{tab:rtops}, most of the transcription
libraries that we have encountered for robotics are 1) written in an OOP
programming style, 2) developed on top of a general-purpose programming
language, and 3) programmed to generate NLPs that are specific to the robot data
and options given.  However, this is at odds with how we express OPs in the
literature.  Mathematically, we do not describe NLPs using OOP constructs and do
not mix the problem formulation with the numerical values of a specific robot.

A unique contribution of Amplify are the design choices behind our library.  The
result is a compact library with fewer dependencies than other trajectory
optimization libraries in robotics.  To be clear, we are not claiming that the
standard approach to ROP design is universally bad and that our approach is
always better.

As stated in Section~\ref{ssec:soc}, we provide an alternative view of what it
means to be modular, reproducible, and lightweight relative to the state of the
art.  This section provides an overview of our design approach, which can be
replicated and improved upon by others.  Overall, our software
\begin{enumerate}[label=D\arabic*]

\item \label{enum:dcl} builds a library on top of a modeling language using a
declarative programming style.

\item \label{enum:soc} separates the problem formulation from the numerical
values that defines a specific instance of the problem,

\item \label{enum:pre} utilizes a presolver to reduce the problem size and tighten bounds,

\item \label{enum:blk} breaks an NLP into self-contained blocks of code, and

\item \label{enum:re} designs the NLP to be reusable and extendable (i.e.,
avoids emitting obfuscated NLP code),

\end{enumerate}
As reference, \ref{enum:dcl}--\ref{enum:pre} are built into the AMPL language.
\ref{enum:blk}--\ref{enum:re} are additional rules we implemented in the design
of Amplify.  In the remainder of this section, we expand on each point in the
subsections below.  Developers can mix and choose which design feature they'd
like to mimic.

\subsection{Let the NLP be Its Own Programming Langauge}
\ref{enum:dcl} presents a different implementation method where we write code in
Amplify at the same level of the decision variables, constraints, and objectives
of an NLP.  We also work in a programming environment, where the syntax of the
language maps more cleanly to the mathematical formulation (e.g., the paper and
pencil definition) of the NLP than a general-purpose programming language.  A
benefit of building on top of an algebraic modeling language is an NLP that is
meant to be edited directly as the contents of the file is the valid syntax of a
Turing-complete programming language.

Furthermore, the declarative nature of the language avoids the levels of
abstraction found in object-oriented programming (OOP) paradigms of other
libraries, which makes editing the NLP more straightforward.  The library itself
is split across 3 files with the core functionality contained in a single file
with 537 lines of source code (LOC) capped at 80 characters per line (740 lines
with comments and blank lines).  The LOC defines the grid points of the NLP, the
constrained hybrid dynamics of the robot, the Runge-Kutta method, and the
B\'{e}zier curves.

Longer term, the declarative programming style also allows for Amplify models to
be an intermediate format for other transcription libraries.  A piece of
software can take in our code and implement the NLP using a pipeline that is
optimized for their uses.  Libraries can also focus on other aspects, like
writing the data file using their own toolchain.

This is effectively how we develop and maintain Amplify.  In our workflow, we
use Mathematica (MMA) to generate the Amplify code.  The MMA code provides
additional conveniences, like a custom parser that simplifies repetitive tasks
(e.g., shorter syntax for defining sets, constraints, and variables), a
topological sorting function that takes in a module and its list of dependencies
and emits only what is needed for the NLP to run in AMPL, and formats data
(e.g., a kinematic tree) in the multi-dimensional table format AMPL expects.
While the MMA code is more complex than the Amplify library itself, the user
does not need our toolchain to run or extend our AMPL code.  A local AMPL
development environment is convenient, but an Internet connection is sufficient
to reproduce the optimal trajectory and a text editor is enough to extend the
emitted NLP.
\begin{table*}[t]
\centering
\scriptsize
\caption{Wall time and problem size comparisons.  Wall time comparison is
between local and NEOS computations across 10 runs.  Local timing is the
duration from when the command script starts until the solver returns control
back to the script.  NEOS wall time includes query for submission template,
waiting in queue, and running on a compute node.  The Ratio of Means is the
ratio of NEOS's average wall time to the average local wall time.  The problem
size comparison lists the number of variables or constraints created by
Amplify's and the user's initial formulation and the reduction after running
AMPL's presolver on the NLP.}
\label{tab:ns}
\begin{tabular}{
l c
c
c
S[table-format=3.3]
c
c
c
c
}
\hline
& & \multicolumn{1}{c}{\textbf{Local}} & \multicolumn{1}{c}{\textbf{NEOS}} & &
\multicolumn{2}{c}{\textbf{\# of Vars.}} & \multicolumn{2}{c}{\textbf{\# of Eq. and Ineq. Constr.}} \\

\textbf{Demo} & \textbf{RK} &
\textbf{Elapsed Time (s)} &
\textbf{Elapsed Time (s)} &
\textbf{Ratio of Means} &
\textbf{Amplify} &
\textbf{Presolve} &
\textbf{Amplify} &
\textbf{Presolve} \\
\hline

acrobot & TRAP 
& $0.217 \pm 0.037$
& $93.448 \pm 74.709$
& 430.237 
& 8779
& 132
& 8824
& 134 \\

cart-pole (force) & TRAP
& $0.234 \pm 0.039$
& $43.404 \pm 56.936$
& 185.170 
& 13169
& 202
& 13234
& 176 \\

cart-pole (time) & TRAP
& $0.198 \pm 0.022$
& $75.756 \pm 86.706$
& 381.833 
& 4388
& 80
& 4413
& 74 \\

moving block & TRAP
& $0.174 \pm 0.014$
& $20.149 \pm 15.103$
& 116.133 
& 6689 
& 116
& 6722
& 88 \\

five-link biped & HS
& $0.486 \pm 0.054$
& $66.036 \pm 49.696$
& 135.931 
& 99109
& 266
& 98877
& 214 \\

five-link biped (phc) & HS
& $0.551 \pm 0.038$
& $40.512 \pm 56.664$
& 73.472 
& 46821
& 326
& 46820
& 302 \\

planar 7-DOF biped & HS
& $33.008 \pm 1.681$
& $60.344 \pm 26.689$
& 1.828 
& 371155
& 2345
& 372492
& 2537 \\

kinematic car & TRAP
& $0.938 \pm 0.074$
& $21.334 \pm 13.781$
& 22.751
& 16623
& 391
& 16624
& 341 \\

spot & COL3
& $205.670 \pm 7.974$
& $340.847 \pm 52.222$
& 1.657 
& 2460541
& 14854
& 2467453
& 14708 \\

spot & RK1
& $218.941 \pm 3.733$
& $508.254 \pm 120.822$
& 2.321 
& 624391
& 3706
& 627703
& 3560 \\

spot & RK4
& $165.644 \pm 5.254$
& $236.365 \pm 38.433$
& 1.427 
& 2460541
& 14836
& 2467453
& 14690 \\

\hline
\end{tabular}
\end{table*}

\subsection{Separate the NLP Model from Its Data}
There are several design principles that AMPL adopts that influences the
implementation of our library.  The most important is \ref{enum:soc}, the
separation of the model (i.e., the problem formulation) from the data (the
numerical values).  Decomposing a library along these separation of concerns
encourages writing models that are generic and reusable \cite{Fourer2004,
Hart2011}.  For Amplify, this is a necessity as the data is a separate entity
that is not known ahead of time.

When coupled with \ref{enum:dcl}, we designed our library with the intention of
implementing complete algorithms, including the parsing of user inputs from a
data file, using only the parameters, decision variables, and constraints in the
model file.  The result is a stand-alone library (\ref{enum:re}) with syntax
that looks like a NLP, but with decision variables and constraints that are
partitioned into generic and reusable blocks of code (\ref{enum:blk}).

In contrast, other libraries emit NLPs that only represent a specific instance
of a more general OP.  In other words, the model and data are mixed together in
the emitted NLP.  This necessarily couples the NLP to the transcription library
that generated it.  For example, several libraries compute a robot's mass matrix
from user data using the CRBA \cite{Hereid2018, Fevre2020}.  The CRBA is
implemented in a general-purpose programming language.  This does provide an
overall level of generality, but the resulting mass matrix is hardcoded into the
emitted NLP.  Any changes to the model or data components of the NLP are then
expected to go through the library's API, so that it can emit a new custom NLP.
In practice, this is a hindrance to reproducible results because of the user's
reliance on the entire software toolchain to run or make any changes to the NLP
through the API.

\subsection{Limitations}
There are several limitations and room for improvement with Amplify.  Areas of
improvement with the current design include reducing the variable and constraint
count with library modules while still preserving generality.  We also shift the
programming burden onto the user when it comes to generating the data and script
files.  In the end, our minimalistic library would benefit from a frontend that
simplifies and automates portions of the NLP the user has to generate.  Our MMA
code provides many of these features, but we would like to port the code to an
open-source language or web frontend.

We also want to look into adopting formulations that have been introduced in the
literature.  For example, the results in \cite{MorenoMartin2022} provide ideas
for increasing the accuracy of our RK module to be consistent with second-order
dynamics equations.  Finally, solid documentation will be key for broader
adoption.

In terms of limitations, we have listed some throughout this paper and our
workarounds.  There are limitations for which we have not been able to find
workarounds.  For example, while AMPL's syntax and interfaces are very strong
positives of the AML's design, there is concern with continued development given
the closed-source nature of the source code, lack of scoping keywords, concise
syntax for common blocks of NLP code (e.g., integration, expression of the
kinematic tree, etc.), and performance limitations of its backend compiler.  It
is not clear if robotics-oriented tools are part of the company's long-term
plans.

A reproducibility limitation is the inability to specify AMPL and solver version
numbers as part of an NLP's problem setup.  For example, after collecting our
timing data, AMPL upgraded to a later version of Ipopt, which required minor
modifications to our code and led to different performance numbers than reported
in this paper.  We decided not to use the updated version of Ipopt in order to
more closely compare with TROPIC, which ships with a fixed version of Ipopt as
part of its library distribution.

Finally, when it comes to performant code, AMPL's backend compiler is currently
a major limitation.  There is growing evidence to support that AMPL's backend
compiler is not performant enough with respect to CasADi's backend.  Given
AMPL's performance relative to CasADi, we anticipate that Amplify + AMPL will be
treated as an easy prototyping or sandbox environment for testing NLP
formulations or connecting solvers to.  Future work would necessitate an
implementation of Amplify with a CasADi backend to better pinpoint the exact
bottlenecks in the pipeline.

\subsection{Avenues of Future Work}
\subsubsection{Use Presolvers for Performant Code}
The role of a presolver is to remove decision variables and constraints that are
trivially solved and tighten the bounds of the remaining constraints.  This
results in a smaller NLP that is sent to a solver.  AMPL's presolver performs
these tasks over a finite number of scans of the NLP model with the default
being 10.  The exact tests and transformations applied to the model can be found
in \cite{Fourer1994}.  We often see a relative reduction of over $90$\% between
what the solver sees after a presolve and the original problem as formulated in
Amplify.  Table~\ref{tab:ns} provides a summary of problem size reduction in
terms of the number of decision variables and a combined count of equality and
inequality constraints across the examples of Section~\ref{sec:ex}.  In terms of
overhead, our timings show that AMPL's presolver does not significantly
contribute to the total elapsed time.  For our example problems the presolver
completed in under a few milliseconds.

As an avenue of future work, we propose a specialized presolver for ROPs.  At
the very least, a functional ROP presolver removes the burden on the developer
and user from having to explicitly formulate performant code as the process is
principled and automated.  With additional effort, incorporating transformations
proposed in \cite{Betts2010} and other works \cite{Hereid2018, Howell2019,
Klemm2025} that provide known performance benefits further enhance the utility
of a presolver in robotics applications.

\subsubsection{Design an Algebraic Modeling Environment for Robotics}
\label{ssec:raml}
We also propose writing an AMPL-inspired language and development environment
that merges the best of existing robotics trajectory optimization libraries and
AMLs.  Basic features such as easier introspection of an NLP and
read-eval-print loops that can easily modify seed values, add or drop
constraints, and fix decision variables are useful productivity boosters and
facilitates profiling and debugging NLP code.

There are also robotics-specific extensions that should be incorporated.  As
mentioned earlier a declarative language can serve as a standard encoding of a
ROP that is interoperable among different libraries similar to the URDF file for
rigid-body dynamics software.  Whereas AMPL does not focus on integrating with
modern robotics software and hardware design, a ROP-focused development
environment would fill in that gap.  It should define interfaces for plugging
into different automatic differentiation and rigid-body dynamics libraries,
presolvers, and optimization solvers relevant to the robotics community.  Future
work should also consider pioneering the creation of modern workflows, like
running on edge computing hardware, incorporating data-driven processes, and
interfacing with learning-based architectures.

\subsubsection{Build Online Services}
The NEOS server is a convenient online prototyping environment for ROPs.
Table~\ref{tab:ns} compares the time it takes to solve a ROP on a local machine
to the cloud-based machines of the NEOS server.  While not fast enough for many
robotics applications of today, we expect these numbers to improve as Internet
and local network speeds increase.  There are also other benefits that we can
take advantage of now given the maturity of the protocols and technology that
make up the NEOS server as a web service.

In the end, we should contribute our own modeling languages and solvers for
others to easily use or deploy our own NEOS-inspired server.  An online service
tailored towards robotics and related fields creates a clear benefit for
benchmarking ROPs.  Most contributions in the field are often the formulation of
the NLP.  An online service provides a tool for running NLPs on common hardware
with access to a wide range of solvers.  This is in contrast to the current
practice of running a ROP on a personal laptop, which makes comparing run times
across machines difficult.  There has also been an increased interest in custom
solvers for robotics applications.  When coupled with the creation of a standard
format for stating ROPs (Section~\ref{ssec:raml}), we can more easily compare
solvers against benchmark problems and rank them with respect to different
metrics of interest.


\section{Conclusion}
\label{sec:con}
In this paper, we demonstrated Amplify.  Amplify is a lightweight nonlinear
programming framework for solving trajectory optimization problems.  The
framework is built to be reproducible and accessible by design, requiring only
an Internet connection and a text editor.  The framework is built on top of the
mathematical programming language AMPL and the web-based,
optimization-as-a-service NEOS server.  The main take-away with respect to
Amplify's design is that the NLP model is the library.  Furthermore, the design
itself is generic enough that others can incorporate our ideas into their
libraries.

Other unique features of the framework's design are 1) the use of a declarative
modeling language to formulate a trajectory optimization problem, 2) the
implementation of efficient rigid-body dynamics algorithms as constraints for
modeling jump maps and acceleration-level constraints of a hybrid trajectory, 3)
the use of Butcher tableaus to specify arbitrary Runge-Kutta methods for solving
the continuous-time dynamics as user inputs, 4) constraints for specifying
B\'{e}zier trajectories and their derivatives, and 5) the use of AMPL's
presolver to reduce the problem size and improve computational performance.  A
comparison with three other libraries are presented along with a discussion of
Amplify's limitation.  We also outline avenues for future work that developers
in the robotics trajectory optimization community should pursue.

\bibliographystyle{IEEEtran}
\bibliography{root}

@Book{Betts2020,
  author           = {Betts, John T.},
  title            = {Practical methods for optimal control using nonlinear programming},
  edition          = {Third edition},
  isbn             = {9781611976199},
  number           = {36},
  pagetotal        = {1733},
  publisher        = {Society for Industrial and Applied Mathematics},
  series           = {Advances in design and control},
  address          = {Philadelphia},
  creationdate     = {2025-10-02T15:36:01},
  modificationdate = {2025-10-02T15:36:36},
  ppn_gvk          = {1732423768},
  year             = {2020},
}

@Article{Bonsignorio2025,
  author           = {Bonsignorio, Fabio and del Pobil, Angel P. and Zereik, Enrica},
  title            = {Towards reproducible robotics research},
  doi              = {10.1038/s42256-025-01114-7},
  issn             = {2522-5839},
  number           = {10},
  pages            = {1591--1592},
  volume           = {7},
  creationdate     = {2025-12-24T08:06:35},
  journal          = {Nature Machine Intelligence},
  modificationdate = {2026-01-12T09:25:02},
  month            = sep,
  publisher        = {Springer Science and Business Media LLC},
  year             = {2025},
}

@InProceedings{Fevre2019,
  author           = {Fevre, Martin and Lin, Hai and Schmiedeler, James P.},
  booktitle        = {2019 IEEE/RSJ International Conference on Intelligent Robots and Systems (IROS)},
  title            = {Stability and Gait Switching of Underactuated Biped Walkers},
  doi              = {10.1109/iros40897.2019.8967673},
  pages            = {2279--2285},
  publisher        = {IEEE},
  creationdate     = {2025-09-07T05:00:25},
  modificationdate = {2025-09-07T05:00:43},
  month            = nov,
  year             = {2019},
}

@Article{Lee2016,
  author           = {Lee, Leng-Feng and Umberger, Brian R.},
  title            = {Generating optimal control simulations of musculoskeletal movement using OpenSim and MATLAB},
  doi              = {10.7717/peerj.1638},
  issn             = {2167-8359},
  pages            = {e1638},
  volume           = {4},
  comment-nr       = {fmincon vs. IPOPT},
  creationdate     = {2026-01-11T16:02:49},
  journal          = {PeerJ},
  modificationdate = {2026-01-11T16:10:21},
  month            = jan,
  publisher        = {PeerJ},
  year             = {2016},
}

@Book{LaValle2009,
  author           = {LaValle, Steven Michael},
  title            = {Planning Algorithms},
  isbn             = {0521862051},
  pagetotal        = {1842},
  publisher        = {Cambridge University Press},
  url              = {https://lavalle.pl/planning/bookbig.pdf},
  accessdate       = {2025-01-28},
  address          = {Cambridge},
  creationdate     = {2025-01-28T07:55:05},
  modificationdate = {2025-01-28T07:57:20},
  ppn_gvk          = {883406209},
  year             = {2009},
}

@Article{Hereid2018,
  author           = {Hereid, Ayonga and Hubicki, Christian M. and Cousineau, Eric A. and Ames, Aaron D.},
  title            = {Dynamic {Humanoid} {Locomotion}: {A} {Scalable} {Formulation} for {HZD} {Gait} {Optimization}},
  doi              = {10.1109/TRO.2017.2783371},
  issn             = {1941-0468},
  number           = {2},
  pages            = {370--387},
  url              = {https://ieeexplore.ieee.org/abstract/document/8260563},
  urldate          = {2024-06-21},
  volume           = {34},
  creationdate     = {2024-06-21T15:39:05},
  journal          = {IEEE Transactions on Robotics},
  modificationdate = {2025-06-14T14:54:03},
  month            = apr,
  shorttitle       = {Dynamic {Humanoid} {Locomotion}},
  year             = {2018},
}

@Article{Kim2023,
  author           = {Kim, Soohwan and Kim, Minkyoung},
  title            = {Rotation Representations and Their Conversions},
  doi              = {10.1109/access.2023.3237864},
  issn             = {2169-3536},
  pages            = {6682--6699},
  volume           = {11},
  creationdate     = {2024-07-17T21:19:04},
  journal          = {IEEE Access},
  modificationdate = {2024-07-17T21:19:32},
  publisher        = {Institute of Electrical and Electronics Engineers (IEEE)},
  year             = {2023},
}

@Article{Wang2024,
  author           = {Wang, Ke and Hu, Zhaoyang Jacopo and Tisnikar, Peter and Helander, Oskar and Chappell, Digby and Kormushev, Petar},
  title            = {When and where to step: Terrain-aware real-time footstep location and timing optimization for bipedal robots},
  doi              = {10.1016/j.robot.2024.104742},
  issn             = {0921-8890},
  pages            = {104742},
  volume           = {179},
  creationdate     = {2025-07-13T13:03:12},
  journal          = {Robotics and Autonomous Systems},
  modificationdate = {2025-07-13T13:03:43},
  month            = sep,
  publisher        = {Elsevier BV},
  year             = {2024},
}

@Article{Jallet2025,
  author           = {Jallet, Wilson and Bambade, Antoine and Arlaud, Etienne and El-Kazdadi, Sarah and Mansard, Nicolas and Carpentier, Justin},
  title            = {ProxDDP: Proximal Constrained Trajectory Optimization},
  doi              = {10.1109/tro.2025.3554437},
  issn             = {1941-0468},
  pages            = {2605--2624},
  volume           = {41},
  creationdate     = {2025-08-19T17:39:55},
  journal          = {IEEE Transactions on Robotics},
  modificationdate = {2025-08-19T17:40:13},
  publisher        = {Institute of Electrical and Electronics Engineers (IEEE)},
  year             = {2025},
}

@Article{Czyzyk1998,
  author           = {Czyzyk, J. and Mesnier, M.P. and More, J.J.},
  title            = {The NEOS Server},
  doi              = {10.1109/99.714603},
  issn             = {1070-9924},
  number           = {3},
  pages            = {68--75},
  volume           = {5},
  creationdate     = {2025-11-08T07:33:51},
  journal          = {IEEE Computational Science and Engineering},
  modificationdate = {2025-11-08T07:35:36},
  publisher        = {Institute of Electrical and Electronics Engineers (IEEE)},
  year             = {1998},
}

@Book{Betts2010,
  author           = {Betts, John T.},
  title            = {Practical Methods for Optimal Control and Estimation Using Nonlinear Programming},
  doi              = {10.1137/1.9780898718577},
  edition          = {Second},
  isbn             = {http://id.crossref.org/isbn/978-0-89871-857-7},
  publisher        = {Society for Industrial \& Applied Mathematics (SIAM)},
  url              = {http://dx.doi.org/10.1137/1.9780898718577},
  creationdate     = {2025-04-17T12:37:35},
  modificationdate = {2026-02-26T11:22:51},
  month            = {Jan},
  year             = {2010},
}

@Article{Czyzyk1999,
  author           = {Czyzyk, Joseph and Wisniewski, Timothy and Wright, Stephen J.},
  title            = {Optimization Case Studies in the NEOS Guide},
  doi              = {10.1137/s0036144598334874},
  issn             = {1095-7200},
  number           = {1},
  pages            = {148--163},
  volume           = {41},
  creationdate     = {2025-11-08T07:33:11},
  journal          = {SIAM Review},
  modificationdate = {2025-11-08T07:33:28},
  month            = jan,
  publisher        = {Society for Industrial & Applied Mathematics (SIAM)},
  year             = {1999},
}

@TechReport{Gay1996,
  author           = {David M. Gay},
  institution      = {AT\&T Bell Laboratories},
  title            = {More AD of Nonlinear AMPL Models: Computing Hessian Information and Exploiting Partial Separability},
  note             = {Technical Report},
  url              = {https://ampl.com/REFS/hess.pdf},
  address          = {Murray Hill, NJ, USA},
  creationdate     = {2025-12-26T12:04:00},
  modificationdate = {2025-12-26T12:05:39},
  year             = {1996},
}

@Misc{MathWorks2025,
  author           = {{The MathWorks, Inc.}},
  title            = {{fmincon — Find minimum of constrained nonlinear multivariable function}},
  howpublished     = {\url{https://www.mathworks.com/help/optim/ug/fmincon.html}},
  note             = {[Accessed: 2025-12-26]},
  url              = {https://www.mathworks.com/help/optim/ug/fmincon.html},
  creationdate     = {2025-12-26T11:38:54},
  modificationdate = {2025-12-26T11:41:48},
  year             = {2025},
}

@Article{Featherstone2010b,
  author           = {Featherstone, Roy},
  title            = {Exploiting {Sparsity} in {Operational}-space {Dynamics}},
  doi              = {10.1177/0278364909357644},
  issn             = {0278-3649},
  language         = {en},
  number           = {10},
  pages            = {1353--1368},
  url              = {https://doi.org/10.1177/0278364909357644},
  urldate          = {2024-09-15},
  volume           = {29},
  creationdate     = {2024-09-15T22:02:18},
  journal          = {The International Journal of Robotics Research},
  modificationdate = {2024-09-15T22:03:29},
  month            = sep,
  publisher        = {SAGE Publications Ltd STM},
  year             = {2010},
}

@Article{Cervera2024,
  author           = {Cervera, Enric},
  title            = {Run to the Source: The Effective Reproducibility of Robotics Code Repositories},
  doi              = {10.1109/mra.2023.3336470},
  issn             = {1558-223X},
  number           = {2},
  pages            = {125--134},
  volume           = {31},
  creationdate     = {2025-12-24T08:15:08},
  journal          = {IEEE Robotics \& Automation Magazine},
  modificationdate = {2025-12-24T08:24:46},
  month            = jun,
  publisher        = {Institute of Electrical and Electronics Engineers (IEEE)},
  year             = {2024},
}

@Article{Dunning2017,
  author           = {Dunning, Iain and Huchette, Joey and Lubin, Miles},
  title            = {JuMP: A Modeling Language for Mathematical Optimization},
  doi              = {10.1137/15m1020575},
  issn             = {1095-7200},
  number           = {2},
  pages            = {295--320},
  volume           = {59},
  creationdate     = {2026-01-29T13:45:56},
  journal          = {SIAM Review},
  modificationdate = {2026-01-29T13:47:30},
  month            = jan,
  publisher        = {Society for Industrial & Applied Mathematics (SIAM)},
  year             = {2017},
}

@InProceedings{Yamamoto2025,
  author           = {Yamamoto, Koya and Taheri, Ehsan and Junkins, John},
  booktitle        = {2025 American Control Conference (ACC)},
  title            = {Comparison of NLP Solvers and Derivative Accuracy for Solving Multi-Impulse Cislunar Trajectory Optimization Problems},
  doi              = {10.23919/acc63710.2025.11107869},
  pages            = {3807--3812},
  publisher        = {IEEE},
  comment-nr       = {fmincon vs. IPOPT},
  creationdate     = {2026-01-11T16:09:09},
  modificationdate = {2026-01-11T16:10:34},
  month            = jul,
  year             = {2025},
}

@Book{Westervelt2007,
  author           = {Westervelt, Eric R. and Jessy W. Grizzle and Christine Chevallereau and Jun Ho Choi and Benjamin Morris},
  title            = {Feedback control of dynamic bipedal robot locomotion},
  isbn             = {9781420053722},
  note             = {Literaturverzeichnis: Seite 479-498},
  number           = {1},
  pagetotal        = {503},
  publisher        = {CRC Press, Taylor \& Francis Group},
  series           = {Control and automation},
  address          = {Boca Raton},
  creationdate     = {2024-06-21T13:05:58},
  modificationdate = {2026-01-24T11:27:30},
  ppn_gvk          = {525454217},
  year             = {2007},
}

@Article{Bauchau2007,
  author           = {Bauchau, Olivier A. and Laulusa, André},
  title            = {Review of Contemporary Approaches for Constraint Enforcement in Multibody Systems},
  doi              = {10.1115/1.2803258},
  issn             = {1555-1423},
  number           = {1},
  volume           = {3},
  creationdate     = {2025-06-03T21:20:29},
  journal          = {Journal of Computational and Nonlinear Dynamics},
  modificationdate = {2025-06-03T21:24:37},
  month            = nov,
  publisher        = {ASME International},
  year             = {2007},
}

@Book{Lubich2006,
  author           = {Lubich, Christian},
  title            = {Geometric Numerical Integration},
  edition          = {2nd ed.},
  editor           = {Gerhard Wanner and Ernst Hairer},
  isbn             = {9783540306665},
  note             = {Description based on publisher supplied metadata and other sources.},
  number           = {v.31},
  pagetotal        = {1660},
  publisher        = {Springer Berlin / Heidelberg},
  series           = {Springer Series in Computational Mathematics Ser.},
  subtitle         = {Structure-Preserving Algorithms for Ordinary Differential Equations},
  address          = {Berlin, Heidelberg},
  creationdate     = {2025-06-14T15:11:57},
  modificationdate = {2025-06-14T15:12:10},
  ppn_gvk          = {174883245X},
  year             = {2006},
}

@Book{Khalil2002,
  author           = {Khalil, Hassan K.},
  title            = {{N}onlinear {S}ystems},
  edition          = {3rd},
  isbn             = {0130673897},
  pagetotal        = {750},
  publisher        = {Prentice Hall},
  subtitle         = {Hassan K.},
  address          = {Upper Saddle River, NJ},
  creationdate     = {2025-08-01T22:46:36},
  modificationdate = {2025-08-24T22:01:43},
  ppn_gvk          = {1169419429},
  year             = {2002},
}

@InProceedings{Balkcom2000,
  author           = {Devin J. Balkcom and Matthew T. Mason},
  booktitle        = {Proceedings of the Fourth Workshop on the Algorithmic Foundations of Robotics (WAFR)},
  title            = {Geometric Construction of Time Optimal Trajectories for Differential Drive Robots},
  note             = {Also appears in “Algorithmic and Computational Robotics: New Directions”, A. K. Peters/CRC Press},
  url              = {https://rlab.cs.dartmouth.edu/publications/djb-wafr00.pdf},
  address          = {Cambridge, MA, USA},
  creationdate     = {2025-12-26T15:30:40},
  modificationdate = {2025-12-26T15:32:14},
  year             = {2000},
}

@Article{Kelly2017,
  author           = {Kelly, Matthew},
  title            = {An Introduction to Trajectory Optimization: How to Do Your Own Direct Collocation},
  doi              = {10.1137/16m1062569},
  issn             = {1095-7200},
  number           = {4},
  pages            = {849--904},
  volume           = {59},
  creationdate     = {2025-07-25T11:22:29},
  journal          = {SIAM Review},
  modificationdate = {2026-04-12T22:47:21},
  month            = jan,
  publisher        = {Society for Industrial \& Applied Mathematics (SIAM)},
  year             = {2017},
}

@InProceedings{Vanroye2023a,
  author           = {Vanroye, Lander and Sathya, Ajay and De Schutter, Joris and Decré, Wilm},
  booktitle        = {2023 IEEE/RSJ International Conference on Intelligent Robots and Systems (IROS)},
  title            = {FATROP: A Fast Constrained Optimal Control Problem Solver for Robot Trajectory Optimization and Control},
  doi              = {10.1109/iros55552.2023.10342336},
  publisher        = {IEEE},
  creationdate     = {2025-10-29T18:18:31},
  modificationdate = {2025-10-29T18:18:50},
  month            = oct,
  year             = {2023},
}

@Article{Bessonnet2005,
  author           = {Bessonnet, G. and Seguin, P. and Sardain, P.},
  title            = {A {Parametric} {Optimization} {Approach} to {Walking} {Pattern} {Synthesis}},
  doi              = {10.1177/0278364905055377},
  issn             = {0278-3649},
  language         = {en},
  number           = {7},
  pages            = {523--536},
  url              = {https://doi.org/10.1177/0278364905055377},
  urldate          = {2024-06-21},
  volume           = {24},
  creationdate     = {2024-06-21T15:40:26},
  journal          = {The International Journal of Robotics Research},
  modificationdate = {2026-02-17T10:31:21},
  month            = jul,
  publisher        = {SAGE Publications Ltd STM},
  year             = {2005},
}

@Book{Fourer2011,
  author           = {Fourer, Robert and Gay, David M. and Kernighan, Brian W.},
  title            = {{AMPL}: A Modeling Language for Mathematical Programming},
  edition          = {2. ed., [repr.]},
  isbn             = {9780534388096},
  pagetotal        = {517},
  publisher        = {Brooks/Cole},
  subtitle         = {A modeling language for mathematical programming},
  address          = {Belmont, CA [u.a.]},
  creationdate     = {2025-06-03T22:31:39},
  modificationdate = {2025-11-07T17:12:17},
  ppn_gvk          = {1607182254},
  year             = {2011},
}

@InProceedings{Fevre2020,
  author           = {Fevre, Martin and Wensing, Patrick M. and Schmiedeler, James P.},
  booktitle        = {2020 IEEE/RSJ International Conference on Intelligent Robots and Systems (IROS)},
  title            = {Rapid Bipedal Gait Optimization in CasADi},
  doi              = {10.1109/iros45743.2020.9341586},
  pages            = {3672--3678},
  publisher        = {IEEE},
  creationdate     = {2025-12-19T12:16:31},
  modificationdate = {2025-12-19T12:17:00},
  month            = oct,
  year             = {2020},
}

@Conference{Rosa2026a,
  author           = {Nelson Rosa},
  booktitle        = {The 6th Conference on Modeling, Estimation and Control (MECC 2026)},
  title            = {{A Case Study Comparing Object-Oriented and Declarative Modeling of Robotics Trajectory Optimization Problems}},
  note             = {To Appear.},
  creationdate     = {2026-04-21T22:48:20},
  modificationdate = {2026-07-23T23:37:43},
  month            = apr,
  year             = {2026},
}

@InProceedings{Carpentier2019,
  author     = {Carpentier, Justin and Saurel, Guilhem and Buondonno, Gabriele and Mirabel, Joseph and Lamiraux, Florent and Stasse, Olivier and Mansard, Nicolas},
  booktitle  = {2019 {IEEE}/{SICE} {International} {Symposium} on {System} {Integration} ({SII})},
  title      = {The {Pinocchio} {C}++ library : {A} fast and flexible implementation of rigid body dynamics algorithms and their analytical derivatives},
  doi        = {10.1109/SII.2019.8700380},
  note       = {ISSN: 2474-2325},
  pages      = {614--619},
  url        = {https://ieeexplore.ieee.org/document/8700380},
  urldate    = {2024-04-24},
  issn       = {2474-2325},
  month      = jan,
  shorttitle = {The {Pinocchio} {C}++ library},
  year       = {2019},
}

@InBook{Andersson2012,
  author           = {Andersson, Joel and Åkesson, Johan and Diehl, Moritz},
  booktitle        = {Recent Advances in Algorithmic Differentiation},
  title            = {CasADi: A Symbolic Package for Automatic Differentiation and Optimal Control},
  doi              = {10.1007/978-3-642-30023-3_27},
  isbn             = {9783642300233},
  pages            = {297--307},
  publisher        = {Springer Berlin Heidelberg},
  creationdate     = {2026-02-17T09:29:13},
  issn             = {1439-7358},
  modificationdate = {2026-02-17T09:29:33},
  year             = {2012},
}

@Book{Bainov1993,
  author           = {Bainov, Drumi and Simeonov, Pavel},
  title            = {Impulsive {Differential} {Equations}: {Periodic} {Solutions} and {Applications}},
  edition          = {1st},
  isbn             = {0582096391},
  language         = {en},
  pagetotal        = {1239},
  publisher        = {CRC Press},
  series           = {Monographs and Surveys in Pure and Applied Mathematics},
  subtitle         = {Periodic Solutions and Applications},
  urldate          = {2024-05-18},
  address          = {London},
  journal          = {Routledge \& CRC Press},
  modificationdate = {2026-09-14T13:46:24},
  ppn_gvk          = {1003334016},
  shorttitle       = {Impulsive {Differential} {Equations}},
  year             = {1993},
}

@Misc{Featherstone2012,
  author           = {Roy Featherstone},
  title            = {{S}patial {V}ector and {R}igid-{B}ody {D}ynamics {S}oftware {V}ersion 2},
  note             = {Last accessed: 9/18/2024.},
  url              = {https://royfeatherstone.org/spatial/v2/},
  creationdate     = {2024-09-18T08:45:24},
  modificationdate = {2026-03-22T22:07:34},
  month            = jun,
  year             = {2012},
}

@InProceedings{Mastalli2020,
  author           = {Mastalli, Carlos and Budhiraja, Rohan and Merkt, Wolfgang and Saurel, Guilhem and Hammoud, Bilal and Naveau, Maximilien and Carpentier, Justin and Righetti, Ludovic and Vijayakumar, Sethu and Mansard, Nicolas},
  booktitle        = {2020 IEEE International Conference on Robotics and Automation (ICRA)},
  title            = {Crocoddyl: An Efficient and Versatile Framework for Multi-Contact Optimal Control},
  doi              = {10.1109/icra40945.2020.9196673},
  pages            = {2536--2542},
  publisher        = {IEEE},
  creationdate     = {2025-08-19T17:35:31},
  modificationdate = {2025-08-19T17:35:52},
  month            = may,
  year             = {2020},
}

@Article{Andersson2018,
  author           = {Andersson, Joel A. E. and Gillis, Joris and Horn, Greg and Rawlings, James B. and Diehl, Moritz},
  title            = {CasADi: a software framework for nonlinear optimization and optimal control},
  doi              = {10.1007/s12532-018-0139-4},
  issn             = {1867-2957},
  number           = {1},
  pages            = {1--36},
  volume           = {11},
  creationdate     = {2026-01-27T15:39:46},
  journal          = {Mathematical Programming Computation},
  modificationdate = {2026-01-27T15:40:01},
  month            = jul,
  publisher        = {Springer Science and Business Media LLC},
  year             = {2018},
}

@InProceedings{Griffin2015a,
  author           = {Griffin, Brent and Grizzle, Jessy},
  booktitle        = {2015 American Control Conference (ACC)},
  title            = {Walking gait optimization for accommodation of unknown terrain height variations},
  doi              = {10.1109/acc.2015.7172087},
  pages            = {4810--4817},
  publisher        = {IEEE},
  creationdate     = {2026-02-17T10:35:51},
  modificationdate = {2026-02-17T10:36:13},
  month            = jul,
  year             = {2015},
}

@Article{Rosa2022a,
  author   = {Rosa, Nelson and Lynch, Kevin M.},
  title    = {A {Topological} {Approach} to {Gait} {Generation} for {Biped} {Robots}},
  doi      = {10.1109/TRO.2021.3094159},
  issn     = {1941-0468},
  number   = {2},
  pages    = {699--718},
  url      = {https://ieeexplore.ieee.org/document/9521970},
  urldate  = {2024-03-14},
  volume   = {38},
  journal  = {IEEE Transactions on Robotics},
  month    = apr,
  year     = {2022},
}

@Article{Cervera2019,
  author           = {Cervera, Enric},
  title            = {Try to Start It! The Challenge of Reusing Code in Robotics Research},
  doi              = {10.1109/lra.2018.2878604},
  issn             = {2377-3774},
  number           = {1},
  pages            = {49--56},
  volume           = {4},
  creationdate     = {2025-12-24T08:18:02},
  journal          = {IEEE Robotics and Automation Letters},
  modificationdate = {2025-12-24T08:18:59},
  month            = jan,
  publisher        = {Institute of Electrical and Electronics Engineers (IEEE)},
  year             = {2019},
}

@Article{Ruscelli2022,
  author           = {Ruscelli, Francesco and Laurenzi, Arturo and Tsagarakis, Nikos G. and Mingo Hoffman, Enrico},
  title            = {Horizon: A Trajectory Optimization Framework for Robotic Systems},
  doi              = {10.3389/frobt.2022.899025},
  issn             = {2296-9144},
  volume           = {9},
  creationdate     = {2025-10-30T18:04:38},
  journal          = {Frontiers in Robotics and AI},
  modificationdate = {2025-10-30T18:04:55},
  month            = jul,
  publisher        = {Frontiers Media SA},
  year             = {2022},
}

@Article{Fourer1990,
  author           = {Robert Fourer and David M. Gay and Brian W. Kernighan},
  title            = {A Modeling Language for Mathematical Programming},
  number           = {5},
  pages            = {519--554},
  volume           = {36},
  creationdate     = {2026-01-21T14:19:43},
  journal          = {Management Science},
  modificationdate = {2026-01-21T14:27:08},
  month            = may,
  year             = {1990},
}

@Article{Ramezani2013,
  author  = {Ramezani, Alireza and Hurst, Jonathan W. and Akbari Hamed, Kaveh and Grizzle, J. W.},
  title   = {Performance {Analysis} and {Feedback} {Control} of {ATRIAS}, {A} {Three}-{Dimensional} {Bipedal} {Robot}},
  doi     = {10.1115/1.4025693},
  issn    = {0022-0434},
  number  = {021012},
  url     = {https://doi.org/10.1115/1.4025693},
  urldate = {2024-04-21},
  volume  = {136},
  journal = {Journal of Dynamic Systems, Measurement, and Control},
  month   = dec,
  year    = {2013},
}

@Misc{Tedrake2019,
  author           = {Russ Tedrake and the Drake Development Team},
  title            = {Drake: Model-based design and verification for robotics},
  url              = {https://drake.mit.edu},
  modificationdate = {2024-06-21T13:21:26},
  year             = {2019},
}

@InBook{Fourer2004,
  author           = {Fourer, Robert and Gay, David M. and Kernighan, Brian W.},
  booktitle        = {Modeling Languages in Mathematical Optimization},
  title            = {Design Principles and New Developments in the AMPL Modeling Language},
  doi              = {10.1007/978-1-4613-0215-5_7},
  isbn             = {9781461302155},
  pages            = {105--135},
  publisher        = {Springer US},
  creationdate     = {2025-11-07T17:30:29},
  issn             = {1384-6485},
  modificationdate = {2025-11-07T17:30:46},
  year             = {2004},
}

@Article{Klemm2025,
  author           = {Klemm, Victor and de Viragh, Yvain and Rohr, David and Siegwart, Roland and Tognon, Marco},
  title            = {Nonsmooth Trajectory Optimization for Wheeled Balancing Robots With Contact Switches and Impacts},
  doi              = {10.1109/tro.2023.3326334},
  issn             = {1941-0468},
  pages            = {497--517},
  volume           = {41},
  creationdate     = {2025-08-12T18:10:44},
  journal          = {IEEE Transactions on Robotics},
  modificationdate = {2025-08-12T18:11:27},
  publisher        = {Institute of Electrical and Electronics Engineers (IEEE)},
  year             = {2025},
}

@InBook{Fourer1994,
  author           = {Fourer, Robert and Gay, David M.},
  booktitle        = {Large Scale Optimization},
  title            = {{L}arge {S}cale {O}ptimization},
  chapter          = {Experience with a Primal Presolve Algorithm},
  doi              = {10.1007/978-1-4613-3632-7_8},
  editor           = {Hager, W.W. and Hearn, D.W. and Pardalos, P.M.},
  isbn             = {9781461336327},
  pages            = {135--154},
  publisher        = {Springer, Boston, MA},
  creationdate     = {2025-08-19T14:22:18},
  modificationdate = {2026-04-17T00:03:07},
  year             = {1994},
}

@Article{Li2025a,
  author           = {Li, He and Wensing, Patrick M.},
  title            = {Cafe-Mpc: A Cascaded-Fidelity Model Predictive Control Framework With Tuning-Free Whole-Body Control},
  doi              = {10.1109/tro.2024.3504132},
  issn             = {1941-0468},
  pages            = {837--856},
  volume           = {41},
  creationdate     = {2026-01-02T18:03:52},
  journal          = {IEEE Transactions on Robotics},
  modificationdate = {2026-01-02T18:04:56},
  publisher        = {Institute of Electrical and Electronics Engineers (IEEE)},
  year             = {2025},
}

@Book{Featherstone2008,
  author           = {Featherstone, Roy},
  title            = {Rigid Body Dynamics Algorithms},
  doi              = {10.1007/978-1-4899-7560-7},
  isbn             = {9780387743141},
  publisher        = {Springer US},
  creationdate     = {2024-06-21T13:25:24},
  modificationdate = {2024-06-21T13:27:32},
  year             = {2008},
}

@Book{Lynch2023,
  author           = {Lynch, Kevin M. and Frank C. Park},
  title            = {Modern Robotics},
  isbn             = {1107156300},
  note             = {"First published 2017. Reprinted with corrections in 2019 (version 9, June 2023)"--Title page verso. - Title from publisher's bibliographic system (viewed on 04 Jun 2024)},
  pagetotal        = {1528},
  publisher        = {Cambridge University Press},
  subtitle         = {Mechanics, planning, and control},
  address          = {Cambridge, UK},
  creationdate     = {2024-09-18T04:59:37},
  modificationdate = {2026-03-23T12:26:21},
  ppn_gvk          = {1899160779},
  year             = {2023},
}

@Article{Wensing2024,
  author   = {Wensing, Patrick M. and Posa, Michael and Hu, Yue and Escande, Adrien and Mansard, Nicolas and Prete, Andrea Del},
  title    = {Optimization-{Based} {Control} for {Dynamic} {Legged} {Robots}},
  doi      = {10.1109/TRO.2023.3324580},
  issn     = {1941-0468},
  pages    = {43--63},
  url      = {https://ieeexplore.ieee.org/abstract/document/10286076},
  urldate  = {2024-04-12},
  volume   = {40},
  journal  = {IEEE Transactions on Robotics},
  year     = {2024},
}

@InProceedings{MorenoMartin2022,
  author           = {Siro Moreno-Martin AND Lluís Ros AND Enric Celaya},
  booktitle        = {Proceedings of Robotics: Science and Systems},
  title            = {{Collocation Methods for Second Order Systems}},
  doi              = {10.15607/RSS.2022.XVIII.038},
  address          = {New York City, NY, USA},
  creationdate     = {2026-09-11T13:25:23},
  modificationdate = {2026-09-11T14:19:38},
  month            = {June},
  year             = {2022},
}

@Article{Hart2011,
  author           = {Hart, William E. and Watson, Jean-Paul and Woodruff, David L.},
  title            = {Pyomo: modeling and solving mathematical programs in Python},
  doi              = {10.1007/s12532-011-0026-8},
  issn             = {1867-2957},
  number           = {3},
  pages            = {219--260},
  volume           = {3},
  creationdate     = {2026-01-29T13:43:16},
  journal          = {Mathematical Programming Computation},
  modificationdate = {2026-01-29T13:43:31},
  month            = aug,
  publisher        = {Springer Science and Business Media LLC},
  year             = {2011},
}

@Software{Danial2021,
  author           = {Albert Danial},
  title            = {cloc: v2.08},
  doi              = {10.5281/zenodo.5760077},
  url              = {https://doi.org/10.5281/zenodo.5760077},
  version          = {v2.08},
  creationdate     = {2026-02-23T18:49:57},
  modificationdate = {2026-02-23T18:51:12},
  month            = dec,
  publisher        = {Zenodo},
  year             = {2021},
}

@InProceedings{Howell2019,
  author           = {Howell, Taylor A. and Jackson, Brian E. and Manchester, Zachary},
  booktitle        = {2019 IEEE/RSJ International Conference on Intelligent Robots and Systems (IROS)},
  title            = {ALTRO: A Fast Solver for Constrained Trajectory Optimization},
  doi              = {10.1109/iros40897.2019.8967788},
  publisher        = {IEEE},
  creationdate     = {2025-08-12T21:15:02},
  modificationdate = {2025-08-12T21:15:17},
  month            = nov,
  year             = {2019},
}

@Book{Murray1994,
  author           = {Murray, Richard M. and Li, Zexiang and Sastry, S. Shankar},
  title            = {A mathematical introduction to robotic manipulation},
  editor           = {Shankar Sastry},
  isbn             = {0849379814},
  note             = {Includes bibliographical references (p. 441-448) and index},
  pagetotal        = {456},
  publisher        = {CRC Press},
  address          = {Boca Raton, Fla},
  creationdate     = {2025-10-02T15:56:46},
  modificationdate = {2025-10-02T15:56:57},
  ppn_gvk          = {128200367},
  year             = {1994},
}

@Article{Gill2002,
  author           = {Gill, Philip E. and Murray, Walter and Saunders, Michael A.},
  title            = {{SNOPT}: An {SQP} Algorithm for Large-Scale Constrained Optimization},
  doi              = {10.1137/s1052623499350013},
  issn             = {1095-7189},
  number           = {4},
  pages            = {979--1006},
  url              = {http://dx.doi.org/10.1137/S1052623499350013},
  volume           = {12},
  creationdate     = {2025-04-17T12:36:36},
  journal          = {SIAM Journal on Optimization},
  modificationdate = {2025-04-17T12:36:51},
  month            = {Jan},
  publisher        = {Society for Industrial \& Applied Mathematics (SIAM)},
  year             = {2002},
}

@Article{Featherstone2010,
  author   = {Featherstone, Roy},
  title    = {A {Beginner}'s {Guide} to 6-{D} {Vectors} ({Part} 2) [{Tutorial}]},
  doi      = {10.1109/MRA.2010.939560},
  issn     = {1558-223X},
  number   = {4},
  pages    = {88--99},
  url      = {https://ieeexplore.ieee.org/document/5663690},
  urldate  = {2024-05-30},
  volume   = {17},
  journal  = {IEEE Robotics \& Automation Magazine},
  month    = dec,
  year     = {2010},
}

@Article{Jusevicius2021,
  author           = {Jusevičius, Vaidas and Oberdieck, Richard and Paulavičius, Remigijus},
  title            = {Experimental Analysis of Algebraic Modelling Languages for Mathematical Optimization},
  doi              = {10.15388/21-infor447},
  issn             = {1822-8844},
  pages            = {283--304},
  creationdate     = {2025-12-26T11:48:02},
  journal          = {Informatica},
  modificationdate = {2025-12-26T11:48:34},
  publisher        = {Vilnius University Press},
  year             = {2021},
}

@Article{Felis2016,
  author           = {Felis, Martin L.},
  title            = {RBDL: an efficient rigid-body dynamics library using recursive algorithms},
  doi              = {10.1007/s10514-016-9574-0},
  issn             = {1573-7527},
  pages            = {1--17},
  url              = {http://dx.doi.org/10.1007/s10514-016-9574-0},
  creationdate     = {2024-09-18T08:40:24},
  journal          = {Autonomous Robots},
  modificationdate = {2024-09-18T08:40:53},
  year             = {2016},
}

@Article{Griffin2015,
  author           = {Griffin, Brent and Grizzle, Jessy},
  title            = {Nonholonomic virtual constraints for dynamic walking},
  doi              = {10.1109/cdc.2015.7402850},
  url              = {http://dx.doi.org/10.1109/CDC.2015.7402850},
  isbn             = {http://id.crossref.org/isbn/978-1-4799-7886-1},
  journal          = {2015 54th IEEE Conference on Decision and Control (CDC)},
  modificationdate = {2025-12-26T15:00:25},
  month            = {Dec},
  publisher        = {Institute of Electrical and Electronics Engineers (IEEE)},
  year             = {2015},
}

@Article{Waechter2005,
  author           = {Wächter, Andreas and Biegler, Lorenz T.},
  title            = {On the implementation of an interior-point filter line-search algorithm for large-scale nonlinear programming},
  doi              = {10.1007/s10107-004-0559-y},
  issn             = {1436-4646},
  number           = {1},
  pages            = {25--57},
  volume           = {106},
  creationdate     = {2025-12-26T11:27:45},
  journal          = {Mathematical Programming},
  modificationdate = {2025-12-26T15:33:30},
  month            = apr,
  publisher        = {Springer Science and Business Media LLC},
  year             = {2005},
}

\appendix

\section{About the Libraries}
\label{sec:lib}
We present a brief overview of OptimTraj and TROPIC.  Both libraries have demos
that we found straightforward to reproduce based on their documentation and
source code.  The libraries depend on having a Matlab installation.  An analysis
of Horizon can be found in \cite{Rosa2026a}. 

\subsection{OptimTraj}
The OptimTraj library is the companion software to \cite{Kelly2017}, which is an
excellent tutorial guide on implementing one's own transcription library using
direct collocation methods.  The library is written in Matlab and uses fmincon
as its only underlying solver.  The use of fmincon requires installation of
Matlab and its Optimization Toolbox.  There are certain demos that require the
Symbolic Toolbox as well.  Overall, the OptimTraj library provides three
collocation methods without support of an external library: trapezoidal,
Hermite-Simpson, and $4^\text{th}$-order Runge-Kutta integration schemes.  The
user has to provide code for the dynamics $\mathcal{H}$ of the system,
objective, constraints, and any derivatives.  OptimTraj then provides wrapper
functions for converting the user's input into valid arguments for fmincon.  The
emphasis of the library is more for teaching purposes.  It does not claim to be
performative or have a fully-featured API interface.

\subsection{TROPIC}
The TROPIC framework does claim to be performative with an API that provides
similar core features to Amplify (modeling, integration, and B\'{e}zier
functions).  It uses an object-oriented programming approach, which is common
with many transcription libraries (see Table~\ref{tab:rtops}).  The emphasis is
to simplify implementation of an NLP for optimal trajectories of hybrid
dynamical systems with physical and virtual holonomic constraints.  The target
application are for designers with biped robots who want to synthesize a
provably stabilizing feedback control law using the hybrid zero dynamics
framework \cite{Westervelt2007}.

TROPIC has interfaces for specifying the model, objective, and its constraints.
The model is generated using the spatial vector library of
\cite{Featherstone2012}, which also heavily influenced the design of Amplify
(e.g., Section~\ref{sec:rbd}).  The library uses the RNEA, CRBA, and TJA of
\cite{Featherstone2012} to compute the equations of motion as a set of
constraints.  These algorithms take in as input the state and control inputs as
CasADi symbolic variables.  The generated constraints can only be used with a
particular robot as the data (e.g., physical parameters) is baked into the
equations.  With the CasADi symbolic engine, TROPIC is able to compute the
analytic gradients of the objective and constraints.  CasADi is bundled and
tightly integrated into TROPIC's core code.

In terms of other features, the library implements trapezoidal and
Hermite-Simpson integration schemes and can only be used with IPOPT as its
solver.  TROPIC also implements its B\'{e}zier polynomials using
Equation~\ref{eq:bern} with first- and second-order derivatives hard coded by
hand in the library.

\end{document}